\pdfoutput=1

\documentclass[10pt,twocolumn,letterpaper]{article}
\PassOptionsToPackage{table}{xcolor}

\ifdefined\BuildReleased \def\BuildMainOnly{}\fi
\ifdefined\BuildMainOnly\else\ifdefined\BuildSuppOnly\else\def\BuildArxiv{}\fi\fi
\ifdefined\BuildReleased \def\BuildCameraReady{}\fi
\ifdefined\BuildArxiv \def\BuildCameraReady{}\fi

\ifdefined\BuildArxiv
\usepackage[pagenumbers]{cvpr}
\else\ifdefined\BuildCameraReady
\usepackage{cvpr}
\else
\usepackage[review]{cvpr}
\fi\fi

\usepackage{multirow}
\usepackage{comment}
\ifdefined\pdfsuppressptexinfo\pdfsuppressptexinfo=-1\fi
\makeatletter
\newtoks\kernel@output
\kernel@output=\output
\let\kernel@outputdblcol\@outputdblcol
\let\kernel@specialoutput\@specialoutput
\newcommand{\UseKernelTwoColumnOutput}{%
  \global\output=\kernel@output
  \global\let\@outputdblcol\kernel@outputdblcol
  \global\let\@specialoutput\kernel@specialoutput}
\makeatother
\usepackage{cuted}

\usepackage{currfile}

\usepackage{caption}

\AtEndPreamble{
    \crefname{appendix}{Sec.}{Secs.}
    \Crefname{appendix}{Section}{Sections}
}

\usepackage{xspace}
\newcommand{\dataset}{Physicl Scenes\xspace}
\newcommand{\instances}{560K\xspace}
\newcommand{\instancesexact}{567,506\xspace}
\newcommand{\assetsover}{115K\xspace}

\newcommand{\assetsexact}{115K\xspace}
\newcommand{\classes}{1.6K\xspace}
\newcommand{\pickplacetestscenes}{43\xspace}
\newcommand{\pickplacetestepisodes}{222\xspace}

\providecommand{\cmarkc}{\textcolor{green!55!black}{\ding{51}}}
\providecommand{\xmarkc}{\textcolor{black!30}{\ding{55}}}
\providecommand{\pmarkc}{\textcolor{black!45}{(\ding{51})}}
\DeclareUnicodeCharacter{03C0}{\ensuremath{\pi}}

\ifdefined\BuildMainOnly\usepackage{xr-hyper}\fi
\ifdefined\BuildSuppOnly\usepackage{xr-hyper}\fi
\definecolor{cvprblue}{rgb}{0.21,0.49,0.74}
\usepackage[pagebackref,breaklinks,colorlinks,allcolors=cvprblue]{hyperref}
\ifdefined\BuildMainOnly\AtEndPreamble{\externaldocument[][nocite]{supp}[supp_submission.pdf]}\fi
\ifdefined\BuildSuppOnly\AtEndPreamble{\externaldocument[][nocite]{main_submission}}\fi
\usepackage{booktabs,graphicx,pifont}
\usepackage{xcolor}
\usepackage[normalem]{ulem}
\def\confName{CVPR}
\def\confYear{2027}

\title{\dataset{}: An Artist-created 3D Dataset for Learning Perception and Action}

\author{Yueh-Cheng Liu \qquad Snehal Jauhri \qquad Matthias Nie{\ss}ner \qquad Angela Dai\\
Technical University of Munich}

\newcommand{\printreferences}{%
    {\small
    \bibliographystyle{ieeenat_fullname}
    \bibliography{main}}}

\begin{document}
\ifdefined\BuildSuppOnly
\clearpage
\UseKernelTwoColumnOutput
\ifdefined\BuildArxiv\else\setcounter{page}{1}\fi
\maketitlesupplementary
\appendix
\setcounter{figure}{0}
\setcounter{table}{0}
\renewcommand{\thefigure}{A\arabic{figure}}
\renewcommand{\thetable}{A\arabic{table}}
\renewcommand{\theHfigure}{A\arabic{figure}}
\renewcommand{\theHtable}{A\arabic{table}}

\section{Scene types}
\label{sec:supp-scene-types}
\Cref{fig:scene-types} shows the number of scenes per type in \dataset.
\begin{figure*}[t]
  \centering
  \includegraphics[width=0.33\textwidth]{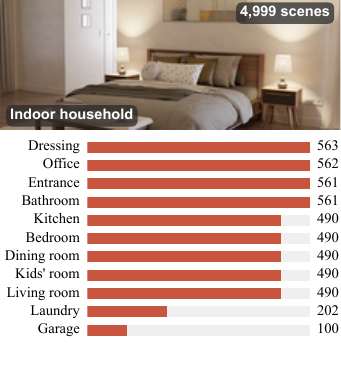}\hfill
  \includegraphics[width=0.33\textwidth]{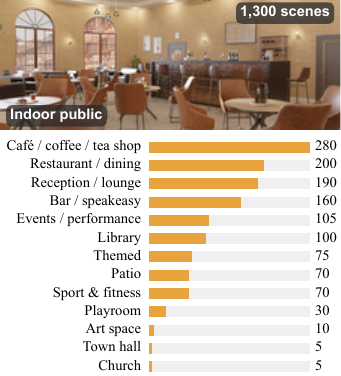}\hfill
  \includegraphics[width=0.33\textwidth]{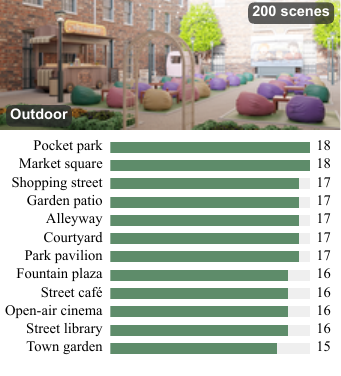}
  \caption{
    \textbf{Scene types in \dataset.}
    Number of scenes per type for indoor household rooms, indoor public venues, and outdoor spaces.}
  \label{fig:scene-types}
\end{figure*}

\section{Dataset creation details}
\label{sec:supp-creation}

\paragraph{Scene authoring.}
Artists compose the scenes from the asset library, starting from a brief for a given type of space, so that layout, object placement, scale, and material use follow real-world logic rather than procedural rules. A reviewer checks each concept against architectural criteria and follows the scene through detailing. Each scene also receives traces of human presence, based on a short scenario of the life that could take place there, so that it does not look artificially clean. A final technical check verifies every scene against the specification. Each scene is stored as a single GLB file with a semantic object hierarchy and PBR textures, in which every object is a separate mesh at metric scale.

\paragraph{Simulation-ready processing.}
We extract the assets and the scene shell (\eg, walls and ceilings) from each scene and compose a scene graph of the physical relations between assets, such as support and attachment.
We classify the assets into categories with a vision-language model (VLM) and have human annotators verify the categorization. To make the scenes physically simulatable, we compute collision hulls for the assets and the scene shell with CoACD~\cite{wei2022approximate}. Using the per-part material annotations of the artist meshes, we assign physical properties (\eg, density) to each part, from which we derive the mass and inertia of each asset. We also use the artists' annotations of emitter materials to define the explicit lighting of each scene, which, combined with the full PBR textures, enables photorealistic rendering. The result is a set of simulation-ready scenes for robotic manipulation and other physics-based tasks.

\section{Dataset comparison details}
\label{sec:supp-comparison}
We compute the statistics in \Cref{tab:table_compare_artist,tab:table_compare_agent} on the data each dataset releases, which can be smaller than what its paper describes.
SpatialGen releases 48 of the 57K rooms in its paper, and GRUtopia releases 99 scenes of a collection that its paper puts at about 100K.

\paragraph{Rooms.}
We count distinct rooms, so a room shipped in several styles counts once.
RoboCasa365 ships 2,500 kitchen scenes that restyle 60 floor plans, so we count 60 rooms and measure the floor area of each plan once.
GRUtopia annotates rooms only in the 30 home scenes of its benchmark, so we count its 321 rooms on those scenes.
For 3D-FRONT, we use the 18,968 furnished rooms reported in its paper, while its floor area covers the whole houses, including unfurnished rooms.
We count each Hypersim scene as one room.
For ease of computation, we treat each \dataset outdoor scene as a single room.

\paragraph{Floor area.}
Floor area is the area of the modeled floor surfaces, summed over all scenes.
For \dataset, the residential, public indoor, and outdoor scenes contribute 162K, 459K, and 2.70M\,m$^2$, respectively.
Outdoor scenes model large ground surfaces, so they account for most of the total. However, even without outdoor scenes, the indoor scenes still cover 621K\,m$^2$, which is larger than the floor area of other datasets in \Cref{tab:table_compare_artist}.

\paragraph{Instances and assets.}
An instance is a placed object, and an asset is a distinct 3D model behind one or more instances.
For 3D-FRONT, we count the placed 3D-FUTURE furniture models, 14,051 in the release, close to the 13,151 reported in its paper.
We exclude the per-house fittings, such as kitchen cabinets, doors, and windows, which carry no object category.

\paragraph{VLM scores.}
We rate the scenes with Qwen3.8-27B, with thinking disabled.
For each dataset with 3D scenes, we sample 20 scenes per room type and render each scene with the same cameras and lighting.
A top-down view from 3.5 times the wall height, with the ceiling removed, shows the layout.
An eye-level view at 1.55\,m with a 100$^\circ$ horizontal field of view, taken from the corner that sees the most furniture, shows the objects.
For datasets that release only images, we use one released frame per scene.
The VLM rates each image from 1 to 5 with the prompts in \Cref{fig:vlm-prompts}, giving its reason before the score, and we report the mean over all rated images.
Layout realism is rated on the top-down view, and asset fidelity and aesthetics on the eye-level view.
Datasets that release only images have no top-down view, so we do not report layout realism for them.

\section{Grasp affordance labels}
\label{sec:supp-affordance}

\Cref{fig:affordance-gallery} shows more objects with affordance labels, in addition to \Cref{fig:grasp-affordance}. Each label marks the part of an object that a grasp should contact, \eg, the handle of a teapot, the frame of a pair of glasses, or the holder of a toilet brush. For example, picking up the glasses by the lenses counts as a success but lowers the affordance accuracy.

\begin{figure*}[t]
  \centering
  \includegraphics[width=\textwidth]{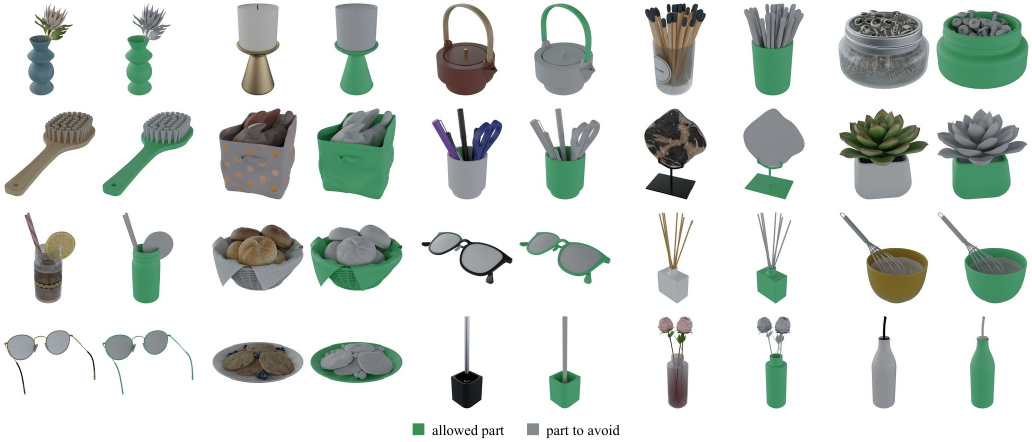}
  \caption{\textbf{Objects with affordance labels in \dataset.}
    Each pair shows an object (left) and its parts (right), with the allowed part that a grasp should contact in green and the other parts in gray.}
  \label{fig:affordance-gallery}
\end{figure*}

\section{Experiment details}
\label{sec:supp-experiments}

\subsection{Inverse rendering}
\label{sec:supp-ir-benchmark}
\paragraph{Benchmark.}
We provide additional benchmarks for single-view and scene-level inverse rendering methods on the \dataset test set, using their released code and weights without fine-tuning. The results are shown in \Cref{tab:inverse-rendering-benchmark}.
We intentionally include both day and night lighting, so that the methods face both sunlight through the windows and scenes lit only by lamps.

For the single-view methods, we take three views per lighting from each of the indoor test scenes, 1,820 frames in total at 640$\times$480.
We skip views that are mostly over- or underexposed, too close to the geometry, or show fewer than three objects.

For the scene-level methods, we take 15 test scenes, one per scene type, 8 with day and 7 with night lighting.
Each scene has 180 input views and 20 held-out views. We query the recovered materials at the surfaces seen by the held-out views, so the scores measure the materials of the whole scene rather than how well a method fits its input views.
Prior scene-level methods, such as FIPT~\cite{wu2023factorized}, IRIS~\cite{lin2025iris}, and Intrinsic Image Fusion~\cite{kocsis2026intrinsic}, evaluate on the four synthetic scenes of FIPT, built from the rendering resources of Bitterli~\cite{bitterli2016resources}, so our scene-level benchmark covers almost four times as many scenes and a much wider range of scene types.
All methods receive the ground-truth camera poses.
FIPT, IRIS, NeILF++, and Intrinsic Image Fusion also assume the ground-truth mesh as input, and we evaluate GenRecon both with and without it. We also use the ground-truth mesh for MVInverse to propagate the prediction to held-out views.

We score albedo in sRGB space, and roughness and metallic in linear space, over the pixels that do not show emitters, windows, or transparent surfaces.
For si-PSNR, we fit one scale per channel in linear space before computing the PSNR.
Even with the ground-truth mesh, the scene-level methods struggle to recover accurate materials, since the complex lighting of the scenes is hard to disentangle from the materials.
For example, the lamps of \dataset are light sources inside the fixtures rather than emissive surfaces, so FIPT finds no emitter at night and bakes the lamp glow into the albedo.
NeILF++, on the other hand, models the lighting as an incident light field that depends on position. We observe that at night, the appearance of the materials is absorbed by the lighting, leaving a mostly uniform albedo.

\begin{table*}[t]
  \centering
  \caption{
    \textbf{Inverse rendering benchmark.} We evaluate the performance of various single-view and scene-level inverse rendering methods without additional fine-tuning on the \dataset test set.
  }
  \label{tab:inverse-rendering-benchmark}
  \begin{tabular}{@{}lcccccc@{}}
    \toprule
    & \multicolumn{4}{c}{Albedo} & Roughness & Metallic \\
    \cmidrule(lr){2-5}\cmidrule(lr){6-6}\cmidrule(lr){7-7}
    Method & si-PSNR $\uparrow$ & PSNR $\uparrow$ & SSIM $\uparrow$ & LPIPS $\downarrow$ & RMSE $\downarrow$ & RMSE $\downarrow$ \\
    \midrule
    \rowcolor{black!8}[0pt][0pt]\multicolumn{7}{@{}l}{\textbf{Single-view methods}} \\
    \quad IntrinsicImageDiffusion~\cite{kocsis2024intrinsic} & 16.39 & 15.20 & 0.709 & 0.361 & 0.279 & 0.252 \\
    \quad Ouroboros~\cite{sun2025ouroboros} & 19.48 & 18.38 & 0.735 & 0.318 & 0.261 & 0.176 \\
    \quad RGB$\leftrightarrow$X~\cite{zeng2024rgb} & 19.87 & \textbf{18.79} & 0.750 & 0.254 & 0.264 & 0.546 \\
    \quad Marigold-IID~\cite{ke2025marigold} & 19.92 & 18.25 & 0.737 & 0.266 & 0.291 & 0.187 \\
    \quad DNF-Intrinsic~\cite{zheng2025dnf} & 20.00 & 18.47 & \textbf{0.779} & \textbf{0.239} & 0.347 & 0.249 \\
    \quad MVInverse~\cite{wu2026mvinverse} & 20.28 & 16.91 & 0.759 & 0.326 & 0.269 & \textbf{0.159} \\
    \quad DiffusionRenderer~\cite{liang2025diffusionrenderer} & \textbf{20.78} & 18.56 & 0.768 & 0.277 & \textbf{0.233} & 0.426 \\
    \midrule
    \rowcolor{black!8}[0pt][0pt]\multicolumn{7}{@{}l}{\textbf{Scene-level methods}} \\
    \quad NeILF++~\cite{zhang2023neilf++} & 14.74 & 9.82 & 0.607 & 0.590 & 0.381 & 0.181 \\
    \quad GS-IR~\cite{liang2024gs} & 15.36 & 14.05 & 0.612 & 0.508 & 0.421 & 0.578 \\
    \quad GenRecon~\cite{schmid2026genrecon} w/o GT geometry & 13.75 & 12.57 & 0.526 & 0.489 & 0.340 & 0.135 \\
    \quad GenRecon~\cite{schmid2026genrecon} w/ GT geometry & 16.95 & 15.03 & 0.650 & 0.364 & 0.335 & \textbf{0.122} \\
    \quad IRIS~\cite{lin2025iris} & 17.00 & 14.81 & 0.706 & 0.365 & 0.400 & 0.139 \\
    \quad FIPT~\cite{wu2023factorized} & 17.75 & 14.81 & 0.702 & \textbf{0.334} & 0.327 & 0.131 \\
    \quad Intrinsic Image Fusion~\cite{kocsis2026intrinsic} & 18.46 & \textbf{15.98} & \textbf{0.734} & 0.343 & 0.385 & 0.142 \\
    \quad MVInverse~\cite{wu2026mvinverse} & \textbf{19.27} & 14.97 & 0.705 & 0.376 & \textbf{0.260} & 0.169 \\
    \bottomrule
  \end{tabular}
\end{table*}

\paragraph{Intrinsic decomposition.}
\Cref{fig:ir-compare-supp} shows more examples of the cross-dataset comparison in \Cref{fig:ir-compare}.
\begin{figure*}[t]
  \centering
  \includegraphics[width=\textwidth]{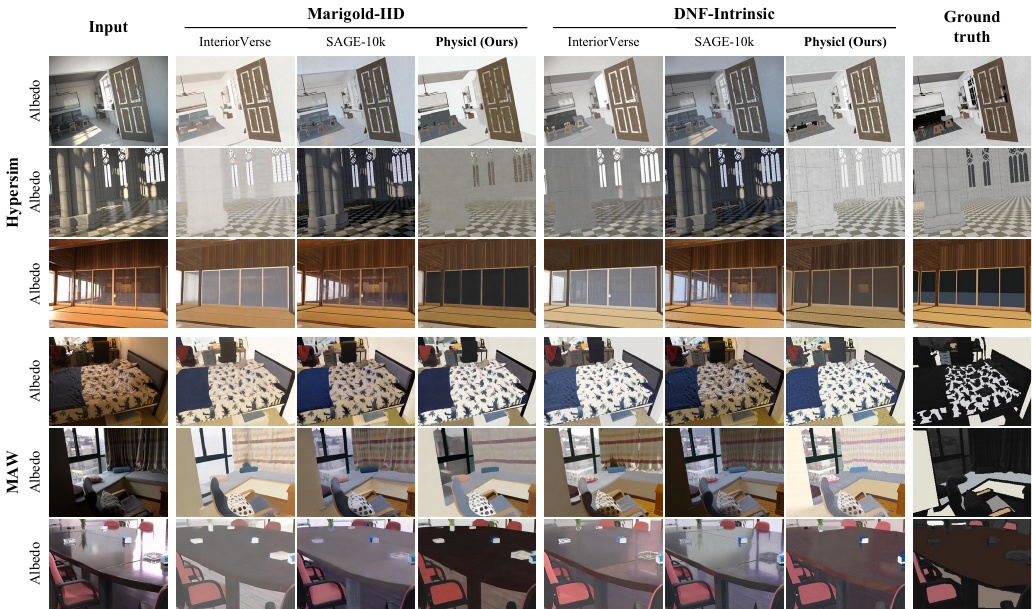}
  \caption{\textbf{Additional qualitative cross-dataset comparison of intrinsic decomposition.}
  Albedo predictions on Hypersim~\cite{roberts2021hypersim} and on the real photos of MAW~\cite{wu2023measured}, in the layout of \Cref{fig:ir-compare}.
  Black ground-truth pixels have no valid label.}
  \label{fig:ir-compare-supp}
\end{figure*}

\subsection{3D instance segmentation}
\label{sec:supp-insseg}
\paragraph{Training data.}
\Cref{fig:insseg-training} shows training scans from \dataset.
For each scene, we back-project the depth maps of the rendered views into one point cloud and keep one point per 1\,cm voxel. Mask3D and PTv3 voxelize this cloud further to 2\,cm.
Each placed object is one instance mask, the pieces of a door or window (\eg, frames, handles) are merged into one instance, and walls, floors, and ceilings carry no instance.
As in real scans, occluded surfaces, such as the floor under furniture, have no points.
\begin{figure*}[t]
  \centering
  \includegraphics[width=\textwidth]{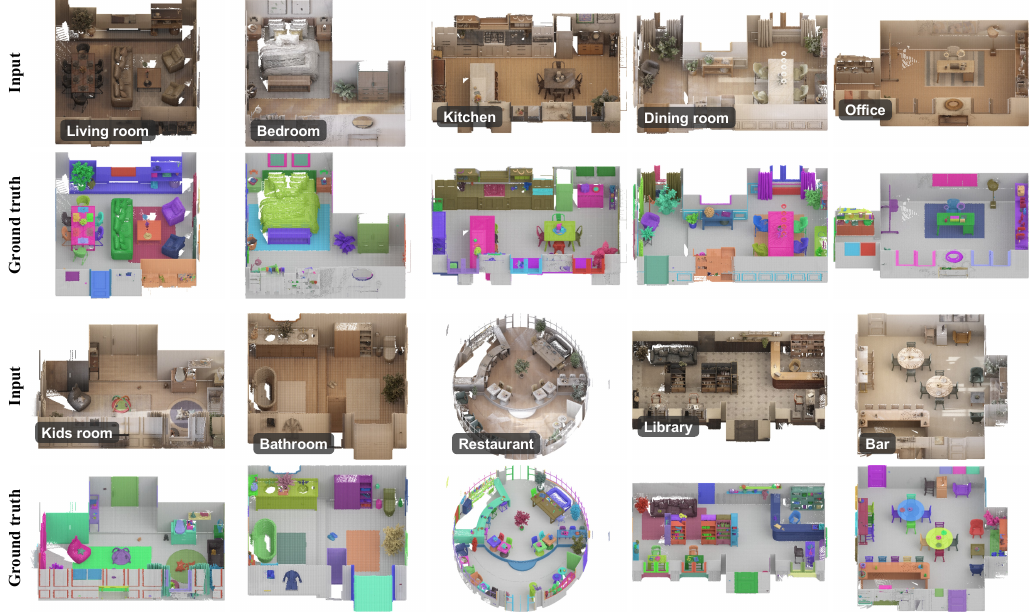}
  \caption{\textbf{Training scans for class-agnostic 3D instance segmentation on \dataset.}
    Each column shows one scene as the back-projected point cloud at the 2\,cm voxel size used in training, in its rendered color (top) and with its ground-truth instance masks (bottom), where walls, floors, and ceilings are gray.
    Points more than 2\,m above the floor are hidden for display.}
  \label{fig:insseg-training}
\end{figure*}

\paragraph{Class-agnostic 3D instance segmentation.}
Each training sample comes from ScanNet200 or the synthetic dataset with equal probability. PTv3 trains for 150k steps with a batch size of 16, and Mask3D for 150k steps with a batch size of 12. We select the checkpoint on the validation sets of the training datasets and never on ScanNet++ or Replica. We train both models class-agnostically. Every instance carries the same label, so the class head of each query only predicts whether the query is an object or no object, instead of one of the 200 ScanNet200 classes.

We evaluate on the 50 official validation scenes of ScanNet++ v2 and on the 8 Replica scenes, taking the vertices of the ground-truth meshes as input points.
The ground-truth instances are the objects of the 84 instance classes of the ScanNet++ top-100 benchmark and of the 48 Replica classes that OpenMask3D and Open-YOLO3D evaluate.
Walls, floors, ceilings, and other structural elements carry no instance.
Both models predict 100 masks per scene at 2\,cm voxels, without DBSCAN, NMS, or a score threshold.
We treat all instances as one class and report the ScanNet AP averaged over IoU thresholds from 0.5 to 0.95, together with AP$_{50}$ and AP$_{25}$.
Instances and predictions with fewer than 100 points are ignored, and a prediction that lies mostly on unlabeled points is not counted as a false positive.

\Cref{fig:seg-compare} shows qualitative class-agnostic instance segmentation results.
\begin{figure*}[t]
  \centering
  \includegraphics[width=\textwidth]{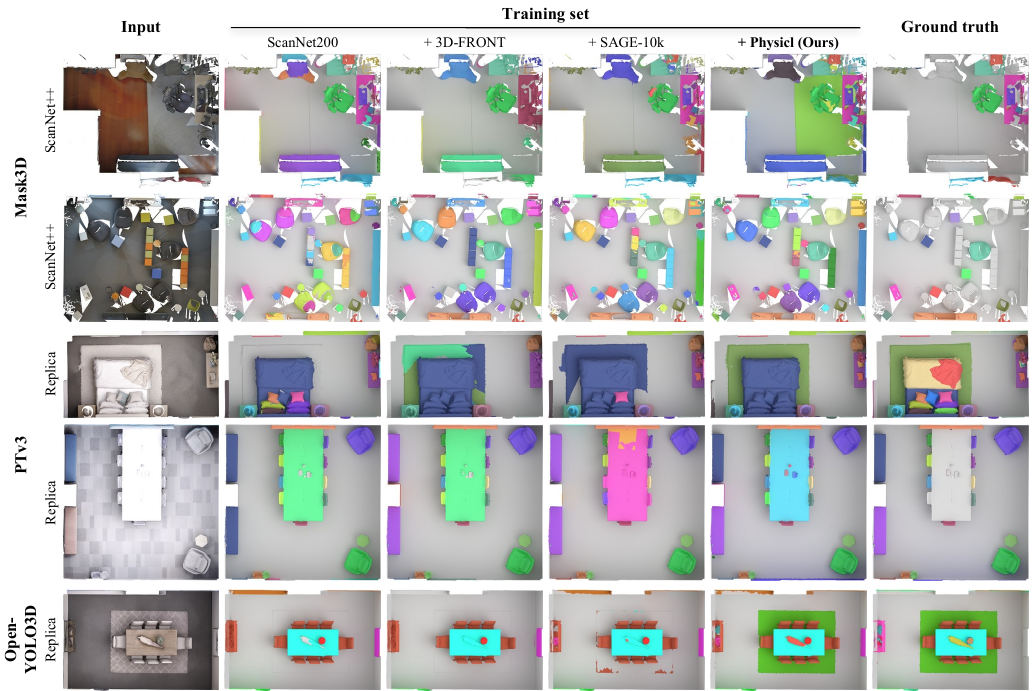}
  \caption{\textbf{Qualitative class-agnostic 3D instance segmentation on ScanNet++ and Replica.}}
  \label{fig:seg-compare}
\end{figure*}

\paragraph{Open-vocabulary 3D instance segmentation.}
We study open-vocabulary instance segmentation with OpenMask3D~\cite{takmaz2023openmask3d} and Open-YOLO3D~\cite{boudjoghra2025open}, using the same training data as the class-agnostic experiments (\Cref{tab:seg-proposals}).
For OpenMask3D and Open-YOLO3D, only the class-agnostic Mask3D proposal network is retrained on each dataset, and the training-free labeling step is unchanged.
We evaluate on the same 8 Replica scenes with the 48 classes and the 200 posed RGB-D frames per scene released by Open-YOLO3D.
Both methods receive the 100 masks per scene of the retrained Mask3D.
We follow the evaluation of both papers and report AP averaged over IoU thresholds from 0.5 to 0.9, and AP$_{50}$.
\par\medskip
\noindent\begin{minipage}{\columnwidth}
  \centering
  \captionof{table}{
    \textbf{Zero-shot open-vocabulary instance segmentation on Replica~\cite{straub2019replica}.}
    We train the class-agnostic Mask3D proposal network for OpenMask3D and Open-YOLO3D with different dataset settings. \dataset and ScanNet200 combined provide the best generalization performance.}
  \label{tab:seg-proposals}
  \setlength{\tabcolsep}{6pt}
  \small
  \resizebox{\ifdim\width>\columnwidth\columnwidth\else\width\fi}{!}{%
  \begin{tabular}{@{}lcc@{}}
    \toprule
    Training data & AP $\uparrow$ & AP$_{50}$ $\uparrow$ \\
    \midrule
    \multicolumn{3}{@{}l}{\textit{OpenMask3D}~\cite{takmaz2023openmask3d}} \\
    ScanNet200 & 11.3 & 16.4 \\
    \quad + 3D-FRONT & 7.5 & 10.3 \\
    \quad + SAGE-10k & 13.5 & 20.6 \\
    \quad + \dataset (Ours) & \textbf{16.6} & \textbf{22.3} \\
    \midrule
    \multicolumn{3}{@{}l}{\textit{Open-YOLO3D}~\cite{boudjoghra2025open}} \\
    ScanNet200 & 20.8 & 27.1 \\
    \quad + 3D-FRONT & 11.8 & 17.5 \\
    \quad + SAGE-10k & 22.6 & 29.9 \\
    \quad + \dataset (Ours) & \textbf{26.1} & \textbf{33.7} \\
    \bottomrule
  \end{tabular}%
  }
\end{minipage}
\par\medskip

\subsection{3D reconstruction}
\label{sec:supp-recon}
\paragraph{Geometry and material reconstruction.}
Besides geometry, GenRecon predicts PBR materials (base color, roughness, and metallic) for each reconstructed scene.
\Cref{fig:genrecon-texture} shows the geometry and albedo that GenRecon, trained on \dataset, reconstructs on held-out test scenes. GenRecon recovers the albedo of the floors, rugs, and furniture under both day and night lighting.
\begin{figure*}[t]
  \centering
  \includegraphics[width=\textwidth]{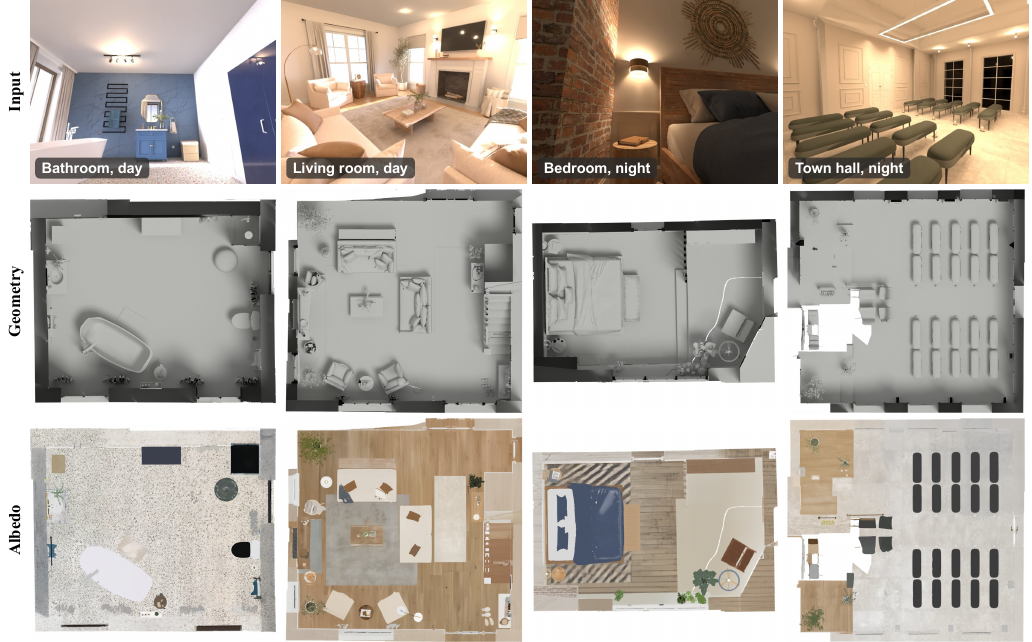}
  \caption{\textbf{Textured reconstructions of GenRecon trained on \dataset.}
    Each column shows one held-out \dataset test scene, one of the input views (top), and the top-down view of the reconstructed geometry (middle) and of the predicted albedo (bottom).
    The reconstructions are cut 1.8\,m above the floor for display.}
  \label{fig:genrecon-texture}
\end{figure*}

\subsection{Mobile manipulation}
\label{sec:supp-manip}
\paragraph{Training details.}
The robot is a Franka arm on a holonomic base, simulated in MuJoCo, and the demonstrations are recorded at 20\,Hz with a wrist camera and a camera on the base, both at 240$\times$320.
The state holds the arm joints, the gripper, and the base velocity.
The action holds the arm joint targets relative to the current joints, an absolute gripper command, and a base target pose relative to the current base pose.

We fully fine-tune all four models from their released checkpoints for 120k steps on four GH200 GPUs and keep an exponential moving average of the weights.
$\pi_{0.5}$ trains with a total batch size of 256 and the other models with 128, depending on the model size and the chunk size.
The action chunk is 50 steps for $\pi_{0.5}$, 40 for GR00T N1.7, 30 for X-VLA, and 16 for MolmoBot, of which the robot executes the first 10 (8 for MolmoBot) before the next query.

\paragraph{Demonstrations.}
\Cref{fig:manip-pnp} shows expert pick-and-place demonstrations in five scenes.
\begin{figure*}[t]
  \centering
  \newcommand{\pnpimg}[1]{\includegraphics[width=0.329\textwidth]{images/manip_pnp/#1.jpg}}
  \newcommand{\pnprow}[2]{%
    \multicolumn{3}{@{}l@{}}{\scriptsize\rule[-2.5pt]{0pt}{10pt}``#2''} \\[1pt]
    \pnpimg{#1_pick} & \pnpimg{#1_move} & \pnpimg{#1_place} \\[2pt]}
  \renewcommand{\arraystretch}{0}
  \setlength{\tabcolsep}{0pt}
  \begin{tabular}{@{}c@{\hspace{0.0065\textwidth}}c@{\hspace{0.0065\textwidth}}c@{}}
    {\footnotesize\textbf{Pick}} & {\footnotesize\textbf{Move}} & {\footnotesize\textbf{Place}} \\[3pt]
    \pnprow{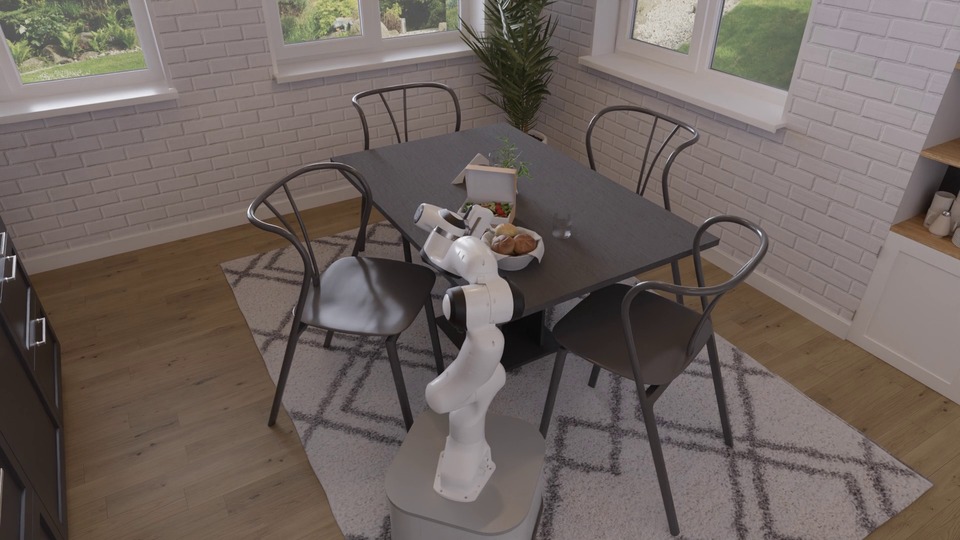}{Pick up the bread basket with cloth and put it on the rectangular pedestal dining table.}
    \pnprow{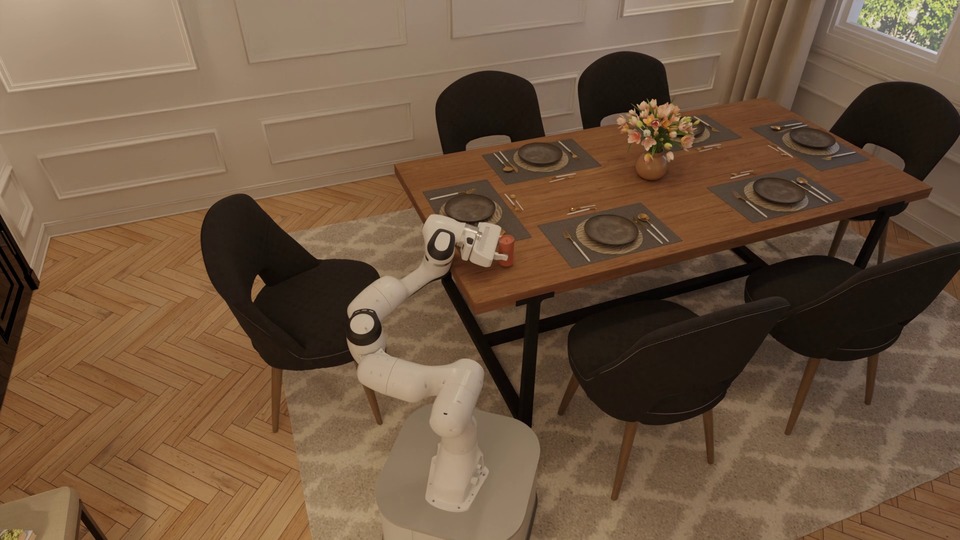}{Put the ceramic flower vase on the wood and metal dining table.}
    \pnprow{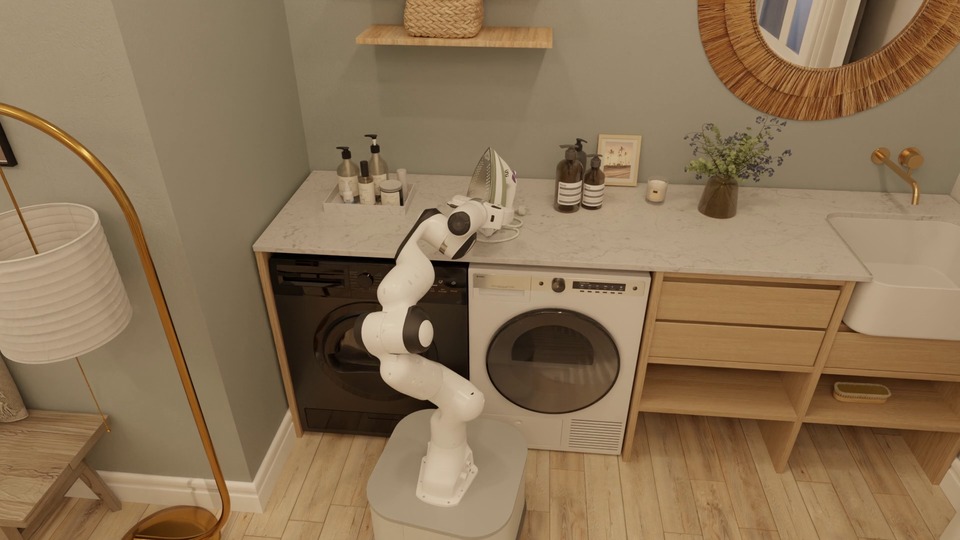}{Go pick up the electric clothes iron, then place it on the wood bathroom vanity.}
    \pnprow{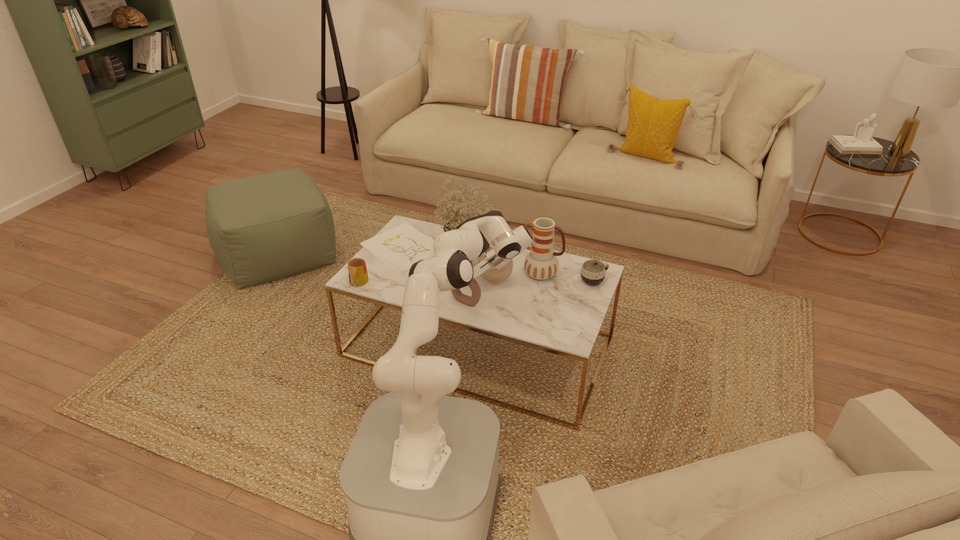}{Go pick up the ceramic loop vase, then place it on the marble top coffee table.}
    \pnprow{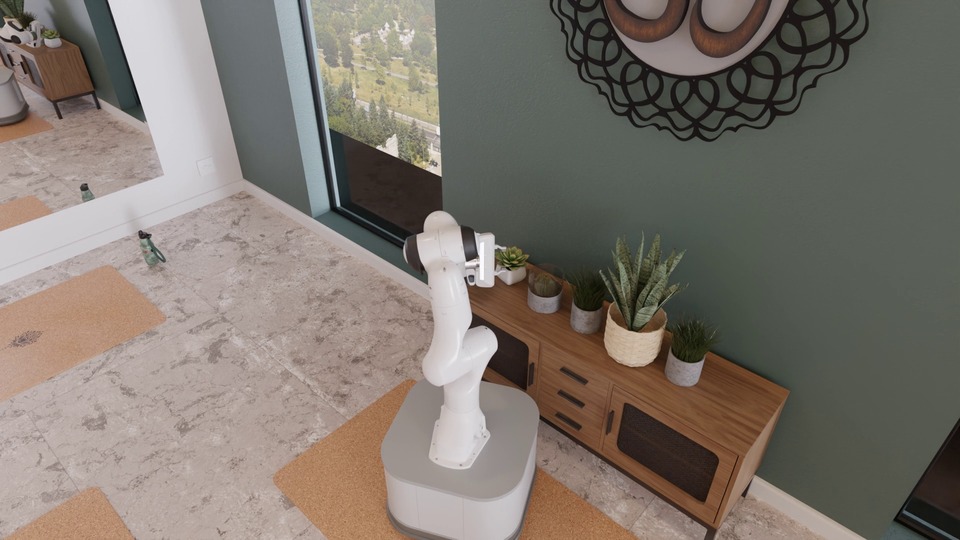}{Pick up the potted artificial succulent and put it on the wooden media sideboard.}
  \end{tabular}
  \caption{
    \textbf{Expert mobile pick-and-place demonstrations in \dataset.}
    Each row shows the instruction of one episode and three frames of its demonstration: the robot lifts the grasped object, carries it across the scene, and releases it on the target.
    The frames are rendered in Blender from the recorded robot and object poses, with a camera that follows the robot.}
  \label{fig:manip-pnp}
\end{figure*}

\section{Monocular depth and normal estimation}
\label{sec:supp-mono}
We additionally provide benchmarks for monocular depth and normal estimation, evaluating existing state-of-the-art models in a zero-shot setting on the \dataset test set (\Crefrange{tab:normal}{tab:depth-metric}).

\paragraph{Setup.}
We use the same test frames as the single-view inverse rendering benchmark (\Cref{sec:supp-ir-benchmark}), 1,820 frames at 640$\times$480, mixture of day and night lighting.
Every method receives the same tone-mapped sRGB image and runs with its released code and weights.
Methods that accept camera intrinsics run both without and with the ground-truth intrinsics (GT $K$), and the other methods only without.
Metric3Dv2 needs the focal length to recover metric scale, so we report its metric depth only with the ground-truth intrinsics.

\paragraph{Metrics.}
We score only pixels with reliable geometry, which excludes glass, mirrors, the sky, surfaces closer than 0.3\,m, and a one-pixel image border, and we score depth only up to 20\,m.
For metric depth (\Cref{tab:depth-metric}), we score the raw prediction.
For relative depth (\Cref{tab:depth-relative}), we fit a scale and a shift per frame by least squares, in depth space for methods that predict depth and in disparity space for methods that predict disparity, and clip the aligned depth to [0.001, 20]\,m.
Methods that predict metric depth also enter the relative table after this alignment.
For normals (\Cref{tab:normal}), we compute the angular error in degrees between the predicted and ground-truth normals in the camera frame, without any alignment.
We compute every metric per frame and average it over all frames of both lightings.
\begin{table*}[t]
  \centering
  \caption{\textbf{Surface normal estimation.}}
  \label{tab:normal}
  \setlength{\tabcolsep}{6pt}
  \begin{tabular}{@{}lccccccc@{}}
    \toprule
    Method & GT $K$ & Mean $\downarrow$ & Median $\downarrow$ & RMSE $\downarrow$ & $11.25^\circ$ $\uparrow$ & $22.5^\circ$ $\uparrow$ & $30^\circ$ $\uparrow$ \\
    \midrule
    \multirow{2}{*}{DSINE~\cite{bae2024rethinking}} & \xmarkc & 13.56 & 6.99 & 22.85 & 70.0 & 82.8 & 86.8 \\
                   & \cmarkc & 13.07 & 6.43 & 22.45 & 71.7 & 83.1 & 87.0 \\
    Lotus-2~\cite{he2025lotus}        & \xmarkc & 12.12 & 7.38 & 19.38 & 71.7 & 86.5 & 90.3 \\
    NormalCrafter~\cite{bin2025normalcrafter}  & \xmarkc & 11.09 & 5.15 & 19.91 & 75.8 & 85.6 & 89.0 \\
    MoGe-2~\cite{wang2025moge}         & \xmarkc & 9.54 & 4.78 & 16.51 & 76.9 & 87.3 & 90.9 \\
    MoGe-3~\cite{kong2026moge}         & \xmarkc & 8.99 & 4.75 & \textbf{15.47} & 78.9 & 88.7 & \textbf{91.9} \\
    Metric3Dv2~\cite{hu2024metric3d}     & \xmarkc & \textbf{8.62} & \textbf{3.59} & 16.99 & \textbf{81.1} & \textbf{88.8} & 91.5 \\
    \bottomrule
  \end{tabular}
\end{table*}

\begin{table*}[t]
  \centering
  \caption{\textbf{Monocular depth estimation (relative depth).}}
  \label{tab:depth-relative}
  \setlength{\tabcolsep}{5pt}
  \resizebox{\ifdim\width>\textwidth\textwidth\else\width\fi}{!}{%
  \begin{tabular}{@{}lccccccccc@{}}
    \toprule
    Method & GT $K$ & AbsRel $\downarrow$ & SqRel $\downarrow$ & RMSE $\downarrow$ & RMSE$_{\log}$ $\downarrow$ & SiLog $\downarrow$ & MAE $\downarrow$ & $\log_{10}$ $\downarrow$ & $\delta_1$ $\uparrow$ \\
    \midrule
    Marigold~\cite{ke2025marigold}       & \xmarkc & 0.049 & 0.014 & 0.155 & 0.075 & 7.45 & 0.103 & 0.021 & 97.7 \\
    Lotus-2~\cite{he2025lotus}        & \xmarkc & 0.044 & 0.028 & 0.182 & 0.064 & 6.36 & 0.113 & 0.019 & 98.4 \\
    DAv2~\cite{yang2024depth}           & \xmarkc & 0.041 & 0.042 & 0.190 & 0.059 & 5.85 & 0.113 & 0.017 & 98.2 \\
    DepthPro~\cite{bochkovskiy2025depth}       & \xmarkc & 0.029 & 0.006 & 0.097 & 0.046 & 4.61 & 0.062 & 0.012 & 99.0 \\
    Metric3Dv2~\cite{hu2024metric3d}     & \xmarkc & 0.026 & 0.007 & 0.100 & 0.050 & 5.00 & 0.054 & 0.011 & 99.0 \\
    \multirow{2}{*}{UniDepthV2~\cite{piccinelli2026unidepthv2}} & \xmarkc & 0.026 & 0.007 & 0.098 & 0.047 & 4.66 & 0.054 & 0.011 & 98.9 \\
                   & \cmarkc & 0.027 & 0.007 & 0.099 & 0.047 & 4.67 & 0.055 & 0.011 & 98.9 \\
    DAv3~\cite{lin2026depth}           & \xmarkc & 0.025 & 0.008 & 0.107 & 0.051 & 5.06 & 0.050 & 0.010 & 98.8 \\
    MoGe-2~\cite{wang2025moge}         & \xmarkc & 0.023 & 0.005 & 0.090 & 0.043 & 4.33 & 0.047 & 0.010 & 99.1 \\
    MoGe-3~\cite{kong2026moge}         & \xmarkc & \textbf{0.017} & \textbf{0.004} & \textbf{0.076} & \textbf{0.036} & \textbf{3.61} & \textbf{0.035} & \textbf{0.008} & \textbf{99.4} \\
    \bottomrule
  \end{tabular}%
  }
\end{table*}

\begin{table*}[t]
  \centering
  \caption{\textbf{Monocular depth estimation (metric depth).}}
  \label{tab:depth-metric}
  \setlength{\tabcolsep}{5pt}
  \resizebox{\ifdim\width>\textwidth\textwidth\else\width\fi}{!}{%
  \begin{tabular}{@{}lccccccccc@{}}
    \toprule
    Method & GT $K$ & AbsRel $\downarrow$ & SqRel $\downarrow$ & RMSE $\downarrow$ & RMSE$_{\log}$ $\downarrow$ & SiLog $\downarrow$ & MAE $\downarrow$ & $\log_{10}$ $\downarrow$ & $\delta_1$ $\uparrow$ \\
    \midrule
    \multirow{2}{*}{UniDepthV2~\cite{piccinelli2026unidepthv2}} & \xmarkc & 0.173 & 0.107 & 0.416 & 0.165 & 5.29 & 0.380 & 0.067 & 78.7 \\
                   & \cmarkc & 0.120 & 0.057 & 0.301 & 0.123 & 5.12 & 0.264 & 0.048 & 91.5 \\
    \multirow{2}{*}{DepthPro~\cite{bochkovskiy2025depth}} & \xmarkc & 0.098 & 0.039 & 0.268 & 0.109 & 5.16 & 0.230 & 0.042 & 92.3 \\
                   & \cmarkc & 0.096 & 0.039 & 0.273 & 0.112 & 5.16 & 0.233 & 0.043 & 91.2 \\
    \multirow{2}{*}{DAv3~\cite{lin2026depth}} & \xmarkc & 0.102 & 0.046 & 0.280 & 0.119 & 5.60 & 0.238 & 0.044 & 90.2 \\
                   & \cmarkc & 0.088 & 0.036 & 0.251 & 0.110 & 5.51 & 0.209 & 0.040 & 91.6 \\
    \multirow{2}{*}{MoGe-2~\cite{wang2025moge}} & \xmarkc & 0.080 & 0.026 & 0.221 & 0.093 & 4.79 & 0.186 & 0.034 & 96.4 \\
                   & \cmarkc & 0.075 & 0.024 & 0.212 & 0.090 & 4.93 & 0.176 & 0.033 & 96.8 \\
    \multirow{2}{*}{Metric3Dv2~\cite{hu2024metric3d}} & \xmarkc & -- & -- & -- & -- & -- & -- & -- & -- \\
                   & \cmarkc & 0.070 & 0.021 & 0.197 & 0.085 & 5.32 & \textbf{0.159} & \textbf{0.030} & \textbf{97.4} \\
    \multirow{2}{*}{MoGe-3~\cite{kong2026moge}} & \xmarkc & 0.072 & 0.022 & 0.202 & 0.084 & \textbf{4.04} & 0.171 & 0.031 & 97.0 \\
                   & \cmarkc & \textbf{0.067} & \textbf{0.020} & \textbf{0.194} & \textbf{0.081} & 4.13 & 0.161 & \textbf{0.030} & 97.0 \\
    \bottomrule
  \end{tabular}%
  }
\end{table*}

\begin{figure*}[t]
  \centering
  \footnotesize
  \fbox{\begin{minipage}{0.97\textwidth}
    \textbf{Shared instructions.}
    You are rating one image of an indoor room. The image shows the room from above or from a point inside it. Seen from above, height is hard to judge: lamps and fans that hang from the ceiling appear on top of the furniture below them, and that is not an error. Rate only what you can see.
    Give a score below 5 only for a specific problem that your reason names. Write your reason first, then the score. Return ONLY a JSON object matching the schema.

    \medskip
    \textbf{Layout realism.}
    Rate how realistic the room's layout is. Look at:
    \begin{itemize}[leftmargin=1.5em,itemsep=0pt,topsep=1pt]
      \item Furniture: is each piece where it would be in a real room, turned the right way for its use, with room to walk and use it?
      \item Small objects and decorations: do lamps, plants, books, tableware, pictures and rugs sit on sensible surfaces and in sensible places, in believable amounts?
      \item Walls, doors and windows: do they form a believable room, and does everything else work with them, for example nothing blocking a door?
    \end{itemize}
    5 = Looks like a real, lived-in room: the furniture, the small objects and decorations, and the walls, doors and windows all make sense.\\
    4 = Realistic, but with one specific oddity. Name it.\\
    3 = The room works, but several things are off.\\
    2 = Clearly wrong in several places: several pieces of furniture face nothing or block doors and paths, several small objects float or sit where they make no sense, or the walls, doors and windows do not form a believable room.\\
    1 = Implausible throughout: many objects overlap or pass through each other or the walls, or most of the placement has no relation to how the room is used.

    \medskip
    \textbf{Asset fidelity.}
    Rate the objects in the room as 3D models: their shape, their materials and their textures. The room is built from 3D models, so compare the objects with other 3D models, not with a photograph. Furniture, fixtures and decor count, built-in furniture included. Walls, floor, ceiling, doors and windows do not.\\
    5 = High-quality 3D models: detailed, accurate shapes and realistic materials and textures (wood grain, fabric, metal, stone) on nearly every object.\\
    4 = Good 3D models, but a few specific objects are clearly simpler or plainly textured. Name them.\\
    3 = Mixed: about half the objects have simplified shapes, or plain, flat or repeated textures.\\
    2 = Mostly crude: most objects are blocky, or carry flat colours or stretched textures.\\
    1 = Broken or unrecognisable: holes, missing parts, untextured blocks, or distorted geometry.

    \medskip
    \textbf{Aesthetics.}
    Rate how good the room looks as a whole: the furniture, materials and colours, and how well they go together. Compare it with other rooms of the same kind.\\
    5 = Well designed: the furniture, colours and materials go together, and the room looks finished and inviting.\\
    4 = Looks good, but with one or two specific weaknesses, for example a plain or generic look. Name them.\\
    3 = Ordinary: acceptable, but plain or unremarkable.\\
    2 = Looks poor: mismatched, too empty or too cluttered, or something looks off.\\
    1 = Looks bad: unappealing, chaotic or broken.
  \end{minipage}}
  \caption{\textbf{VLM prompts for the scores in \Cref{tab:table_compare_artist,tab:table_compare_agent}.}
    Each query holds one image, the shared instructions, one rating question, and a JSON schema with a reason field and a score field.}
  \label{fig:vlm-prompts}
\end{figure*}

\clearpage
\printreferences
\else
\maketitle
\begin{strip}
  \centering
  \includegraphics[width=\textwidth]{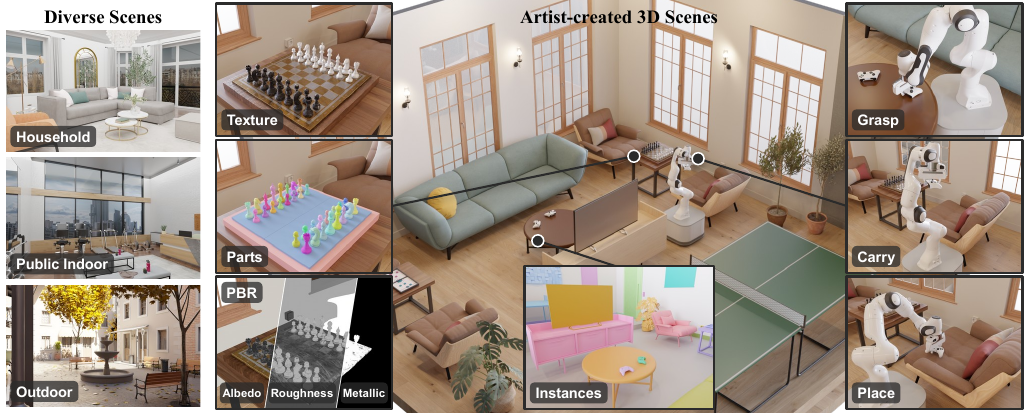}
  \captionsetup{hypcap=false}
  \captionof{figure}{We introduce \textbf{\dataset}: 6.5K artist-created indoor and outdoor scenes with full PBR textures, part annotations, and simulation-ready assets, enabling perception and embodied AI research.}
  \captionsetup{hypcap=true}
  \label{fig:overview}
\end{strip}

\begin{figure*}[t]
  \centering
  \includegraphics[width=\textwidth]{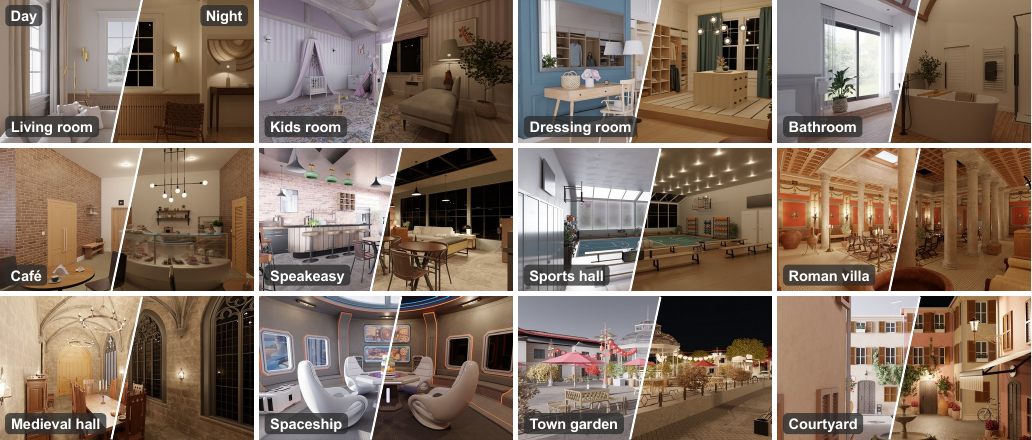}
  \caption{
    \textbf{\dataset{} scenes, shown under day and night lighting settings.}
    \dataset{} covers a diverse set of scenes across different environments spanning household rooms, public indoor venues, themed venues, and outdoor spaces. The explicit lighting defined in the scenes allows for representing the same scene under different illumination conditions.
    }
  \label{fig:scene-gallery}
\end{figure*}

\begin{abstract}

We present \dataset, a high-quality artist-created and curated dataset of 6,499 3D scenes spanning household rooms, public interiors, and outdoor spaces.
The dataset combines detailed geometry, physically based materials, and carefully composed environments with object- and part-level semantic annotations, providing rich, joint supervision for visual and embodied learning.
We demonstrate its value as training data across inverse rendering, 3D reconstruction, scene understanding, and mobile manipulation.
Across these tasks, training state-of-the-art methods with our scenes improves performance over alternative synthetic datasets.
In addition, we introduce a mobile manipulation benchmark, evaluating vision-language-action models performing physical pick and place in diverse 3D scenes with realistic layouts, and using our part annotations to also score where each object is grasped.
Our dataset will be released publicly to facilitate research in visual and embodied AI.
\end{abstract}

\section{Introduction}
\label{sec:intro}

Realistic 3D scenes with fine-grained ground-truth annotations are essential for learning perception and action that generalize reliably to real-world settings.
Perception models for inverse rendering, reconstruction, and scene understanding need dense ground truth, \eg, complete geometry, materials, and instance labels. Embodied agents need interactive environments that behave like the real world, so that navigation and manipulation can be both trained and evaluated in closed-loop settings.

Real-world scans \cite{dai2017scannet,yeshwanth2023scannet++,baruch2021arkitscenes,chang2017matterport3d,ramakrishnan2021habitat} capture real appearance and layouts, but they are noisy and incomplete, and they cannot provide ground truth for albedo, roughness, lighting, unobserved geometry, or object parts.
Synthetic scenes provide all of these by construction, but making them realistic at scale is hard.
Procedural \cite{deitke2022procthor,raistrick2024infinigen} and agentic \cite{yang2024holodeck,pfaff2026scenesmith,xia2026sage} methods scale well, but their realism is bounded by hand-crafted rules or by the quality of generated 3D assets \cite{xiang2025structured,xiang2026native,chen2026sam}.
Artist-created scenes \cite{fu20213d,roberts2021hypersim,zhu2022learning,kolve2017ai2,deitke2020robothor,khanna2024habitat,wang2024grutopia,nasiriany2024robocasa} are more realistic, but manual modeling limits their scale and scene types.

Learning actions adds further requirements.
Scenes must be simulation-ready, with metric scale and physical properties, and their layouts must be realistic enough for navigation.
Part-level annotations also help an agent learn where to interact with an object.
Simulation-ready artist-created scenes exist \cite{kolve2017ai2,khanna2024habitat,wang2024grutopia,nasiriany2026robocasa365} but are limited in scale, and few datasets provide part-level annotations (\Cref{tab:table_compare_artist,tab:table_compare_agent}).
As a result, realistic scenes at scale that are also simulation-ready and annotated at the part level remain scarce.

\providecommand{\cmarkc}{\textcolor{green!55!black}{\ding{51}}}
\providecommand{\xmarkc}{\textcolor{black!30}{\ding{55}}}
\providecommand{\pmarkc}{\textcolor{black!45}{(\ding{51})}}
\providecommand{\outdoorhl}[1]{\textcolor{orange!90!black}{\textbf{#1}}}
\begin{table*}[t]
  \centering
  \caption{\textbf{Comparison with existing artist-created datasets.}
  \dataset is larger in total floor area and in the number of assets and instances.
    SpatialGen and GRUtopia numbers are computed on their released data.
    A VLM scores scenes from 1 to 5 on renderings, and layout realism needs 3D scenes for top-down renders (see \Cref{sec:supp-comparison} in the supplementary material for details).}
  \label{tab:table_compare_artist}
  \setlength{\tabcolsep}{5pt}
  \renewcommand{\arraystretch}{1.15}
  \small
  \resizebox{\ifdim\width>\textwidth\textwidth\else\width\fi}{!}{%
  \begin{tabular}{@{}lrrccccrrcccc@{}}
    \toprule
     & \multicolumn{8}{c}{Property} & \multicolumn{3}{c}{VLM score} & \\
    \cmidrule(lr){2-9}\cmidrule(lr){10-12}
    Dataset & \#Rooms & Floor area (m$^2$) & PBR & 3D & Part & \#Classes & \#Instances & \#Assets & Layout realism $\uparrow$ & Asset fidelity $\uparrow$ & Aesthetic $\uparrow$ & \smash{\shortstack{Mobile\\manipulation}} \\
    \midrule
    SpatialGen~\cite{fang2026spatialgen} & 48 & 848 & \xmarkc & \xmarkc & \xmarkc & 36 & 977 & -- & -- & 3.92 & 4.73 & \xmarkc \\
    Hypersim~\cite{roberts2021hypersim} & 461 & 136.0K & \xmarkc & \pmarkc & \xmarkc & 40 & 63,029 & -- & -- & 3.84 & 4.20 & \xmarkc \\
    InteriorVerse~\cite{zhu2022learning} & -- & -- & \cmarkc & \xmarkc & \xmarkc & -- & -- & -- & -- & 3.11 & 4.46 & \xmarkc \\
    3D-FRONT~\cite{fu20213d} & 18,968 & 578.5K & \xmarkc & \cmarkc & \xmarkc & 49 & 176,826 & 14,051 & 2.72 & 2.54 & 2.86 & \xmarkc \\
    RoboCasa365~\cite{nasiriany2026robocasa365} & 60 & 1.9K & \xmarkc & \cmarkc & \xmarkc & 210 & -- & 3,430 & 2.00 & 2.85 & 3.50 & \cmarkc \\
    RoboTHOR~\cite{deitke2020robothor} & 89 & 4.1K & \cmarkc & \cmarkc & \pmarkc & 47 & 2,883 & 652 & 2.07 & 2.16 & 2.68 & \pmarkc \\
    HSSD-200~\cite{khanna2024habitat} & 2,351 & 59.4K & \cmarkc & \cmarkc & \xmarkc & 466 & 53,021 & 18,656 & 3.09 & 3.33 & 3.62 & \cmarkc \\
    iTHOR~\cite{kolve2017ai2} & 120 & 3.3K & \cmarkc & \cmarkc & \pmarkc & 99 & 3,884 & 2,229 & 2.15 & 2.19 & 2.44 & \cmarkc \\
    GRUtopia~\cite{wang2024grutopia} & 321 & 12.4K & \cmarkc & \cmarkc & \cmarkc & 78 & 37,601 & 34,462 & 2.63 & 4.03 & 4.08 & \cmarkc \\
    \multicolumn{1}{@{}>{\columncolor{yellow!22}[0pt][\tabcolsep]}l}{\textbf{\dataset (Ours)}} & \cellcolor{yellow!22}6,499 & \cellcolor{yellow!22}3.32M & \cellcolor{yellow!22}\cmarkc & \cellcolor{yellow!22}\cmarkc & \cellcolor{yellow!22}\cmarkc & \cellcolor{yellow!22}1,600 & \cellcolor{yellow!22}\instancesexact & \cellcolor{yellow!22}\assetsexact & \cellcolor{yellow!22}\textbf{3.83} & \cellcolor{yellow!22}\textbf{4.54} & \cellcolor{yellow!22}\textbf{4.74} & \multicolumn{1}{>{\columncolor{yellow!22}[\tabcolsep][0pt]}c@{}}{\cmarkc} \\
    \bottomrule
  \end{tabular}%
  }
\end{table*}

We introduce \textbf{\dataset{}}, a large-scale dataset of 6.5K artist-created 3D scenes that combines high visual fidelity, simulation-ready assets, and realistic layouts (see \Cref{fig:overview}). It covers a wide variety of indoor and outdoor scene types. All the scenes contain full PBR textures and rich, explicit lighting, enabling the creation of photorealistic renderings for various downstream applications (see \Cref{fig:scene-gallery}).
\dataset contains \classes object categories and \assetsexact assets, several times more than any existing artist-created dataset (\Cref{tab:table_compare_artist}), and has more realistic layouts, higher asset fidelity, and part-level annotations that procedurally and agentically generated datasets mostly lack (\Cref{tab:table_compare_agent}).

To demonstrate the benefit of \dataset for perception, we train state-of-the-art models for inverse rendering, 3D reconstruction, and 3D instance segmentation, and evaluate them zero-shot on unseen synthetic and real-world datasets.
Compared to prior synthetic datasets, \dataset reduces albedo error on real photographs by up to 47\% and reconstruction Chamfer distance by up to 39\%, and improves 3D instance segmentation AP by up to 5.9 points.

For action, we build a mobile manipulation benchmark in which vision-language-action (VLA) models \cite{intelligence2025pi_,bjorck2025gr00t,zheng2026x,deshpande2026molmob0t} follow language instructions to pick and place objects in complex indoor scenes, rather than on a single tabletop. Success also depends on where an object is grasped: a vase of flowers should be held by the vase, not a stem (\Cref{fig:grasp-affordance}). Using our part annotations, we generate ground-truth trajectories whose grasps respect such affordances, and we measure affordance accuracy, the fraction of successful grasps that contact the part the instruction implies. VLA models trained on \dataset transfer better to real-to-sim ScanNet++ environments than those trained on MolmoSpaces (12.8\% vs.\ 9.5\% success rate).
Training with affordance-aware grasps improves affordance accuracy by 8.7 and 13.5 points and success rate by 1.6 and 5.4 points for $\pi_{0.5}$ and MolmoBot, respectively.

\noindent We summarize the contributions below:
\begin{itemize}
    \item
    We introduce \dataset, 6.5K artist-created, simulation-ready 3D scenes with PBR textures, realistic layouts, and part-level annotations, covering indoor and outdoor spaces.
    \item We show that training on \dataset improves zero-shot generalization to unseen and real-world data for vision tasks like inverse rendering, 3D reconstruction, and 3D instance segmentation over prior synthetic datasets.
    \item We introduce a mobile manipulation benchmark with ground-truth trajectories in realistic indoor scenes. We add an affordance accuracy metric and affordance-aware grasp supervision derived from our part annotations. VLA models trained on \dataset transfer better to real-to-sim scenes.
\end{itemize}

\section{Related work}
\providecommand{\cmarkc}{\textcolor{green!55!black}{\ding{51}}}
\providecommand{\xmarkc}{\textcolor{black!30}{\ding{55}}}
\providecommand{\pmarkc}{\textcolor{black!45}{(\ding{51})}}
\providecommand{\outdoorhl}[1]{\textcolor{orange!90!black}{\textbf{#1}}}
\begin{table}[t]
  \centering
  \caption{\textbf{Comparison with existing procedurally and agentically generated datasets.} \dataset contains higher-fidelity assets, more realistic layouts, and better aesthetic quality.
  }
  \label{tab:table_compare_agent}
  \setlength{\tabcolsep}{2.5pt}
  \renewcommand{\arraystretch}{1.15}
  \footnotesize
  \resizebox{\ifdim\width>\columnwidth\columnwidth\else\width\fi}{!}{%
  \begin{tabular}{@{}lcccccccc@{}}
    \toprule
     & \multicolumn{4}{c}{Property} & \multicolumn{3}{c}{VLM score $\uparrow$} & \\
    \cmidrule(lr){2-5}\cmidrule(lr){6-8}
    Dataset & PBR & Lighting & 3D & Part & \shortstack{Layout\\realism} & \shortstack{Asset\\fidelity} & Aesthetic & \shortstack{Mobile\\manip.} \\
    \midrule
    Aria Synthetic~\cite{avetisyan2024scenescript} & \xmarkc & render-only & \xmarkc & \xmarkc & -- & 2.01 & 2.07 & \xmarkc \\
    Holodeck~\cite{yang2024holodeck} & \pmarkc & explicit & \cmarkc & \xmarkc & 2.05 & 2.44 & 2.15 & \pmarkc \\
    MolmoSpaces~\cite{kim2026molmospaces} & \xmarkc & explicit & \cmarkc & \xmarkc & 1.66 & 2.34 & 2.27 & \cmarkc \\
    InternScenes~\cite{zhong2025internscenes} & \pmarkc & explicit & \cmarkc & \xmarkc & 1.63 & 2.38 & 2.34 & \cmarkc \\
    SAGE~\cite{xia2026sage} & \pmarkc & none & \cmarkc & \xmarkc & 1.53 & 2.95 & 2.61 & \cmarkc \\
    SceneSmith~\cite{pfaff2026scenesmith} & \pmarkc & none & \cmarkc & \xmarkc & 2.90 & 2.57 & 3.75 & \cmarkc \\
    ProcTHOR~\cite{deitke2022procthor} & \cmarkc & explicit & \cmarkc & \xmarkc & 1.75 & 2.30 & 2.42 & \cmarkc \\
    Infinigen Indoors~\cite{raistrick2024infinigen} & \cmarkc & explicit & \cmarkc & \pmarkc & 2.55 & 2.25 & 2.44 & \pmarkc \\
    \multicolumn{1}{@{}>{\columncolor{yellow!22}[0pt][\tabcolsep]}l}{\textbf{\dataset (Ours)}} & \cellcolor{yellow!22}\cmarkc & \cellcolor{yellow!22}explicit & \cellcolor{yellow!22}\cmarkc & \cellcolor{yellow!22}\cmarkc & \cellcolor{yellow!22}\textbf{3.83} & \cellcolor{yellow!22}\textbf{4.54} & \cellcolor{yellow!22}\textbf{4.74} & \multicolumn{1}{>{\columncolor{yellow!22}[\tabcolsep][0pt]}c@{}}{\cmarkc} \\
    \bottomrule
  \end{tabular}%
  }
\end{table}

\subsection{3D datasets}

Real-world 3D datasets are captured with handheld RGB-D sensors~\cite{dai2017scannet,hua2016scenenn,straub2019replica}, tripod-mounted panoramic RGB-D cameras~\cite{chang2017matterport3d,armeni2017joint,xia2018gibson,ramakrishnan2021habitat}, or stationary laser scanners~\cite{baruch2021arkitscenes,yeshwanth2023scannet++}.
Despite their realism, most scans are noisy and miss occluded regions.
Artist-made synthetic datasets, on the other hand, avoid scanning artifacts and provide exact ground truth.
Some release rendered images of professionally modeled interiors~\cite{roberts2021hypersim,zhu2022learning,fang2026spatialgen}, and others release 3D scenes for scene understanding and embodied agents~\cite{fu20213d,kolve2017ai2,deitke2020robothor,khanna2024habitat,wang2024grutopia,nasiriany2024robocasa,nasiriany2026robocasa365}.
Both are limited in scale and scene types by the effort of manual modeling.
To overcome this, procedural methods~\cite{deitke2022procthor,avetisyan2024scenescript,raistrick2024infinigen,kim2026molmospaces,zhong2025internscenes} sample layouts and assets from rules or constraints, and agentic methods~\cite{yang2024holodeck,xia2026sage,pfaff2026scenesmith} compose scenes with LLMs from existing or generated 3D assets~\cite{xiang2025structured,xiang2026native,chen2026sam,zhao2025hunyuan3d,xiang2020sapien,deitke2023objaverse}.
However, their realism is bounded by their generation rules and asset quality.
In comparison, \dataset is more realistic than both prior artist-made and generated datasets, and larger and more diverse than the artist-made ones, with 6.5K scenes spanning household rooms, public venues, and outdoor spaces (\Cref{tab:table_compare_artist,tab:table_compare_agent}).

\subsection{Navigation and manipulation benchmarks}
Many simulated benchmarks for embodied agents focus either on navigation~\cite{savva2019habitat,shen2021igibson,batra2020objectnav,krantz2020beyond,khanna2024goat} or on tabletop manipulation with a fixed arm~\cite{liu2023libero,fei2025libero,mees2022calvin,james2020rlbench,tao2025demonstrating,li2024evaluating,zhang2025vlabench,han2025robocerebra,gao2025genmanip}, which has become the standard testbed for VLA models~\cite{black2024pi_0,intelligence2025pi_,bjorck2025gr00t,zheng2026x,deshpande2026molmob0t}.
Existing mobile manipulation benchmarks combine the two, but often simplify one of them to keep the task tractable.
Some set the place target right next to the object~\cite{nasiriany2024robocasa,kim2026molmospaces,gao2026ebench}, so the base barely moves, while others let the robot travel farther but rely on oracle navigation, magnetic grasps, or assisted grasps~\cite{ehsani2021manipulathor,szot2021habitat,puig2024habitat,yenamandra2023homerobot,shukla2025maniskill,wang2024grutopia,ehsani2024spoc,li2023behavior}.
RoboCasa365~\cite{nasiriany2026robocasa365} comes closest to ours, with physical grasps and long-distance carries, while focusing on kitchen scenes and objects.
In comparison, our benchmark samples place targets on any other receptacle in the scene, often meters from the pick, and evaluates physical grasping and placing in held-out scenes across diverse room types.
The part annotations of \dataset further enable an affordance metric that credits a grasp only when it contacts the part the instruction implies.

\section{\dataset{}}

\subsection{Dataset overview}

\begin{figure*}[t]
  \centering
  \includegraphics[width=0.9\textwidth]{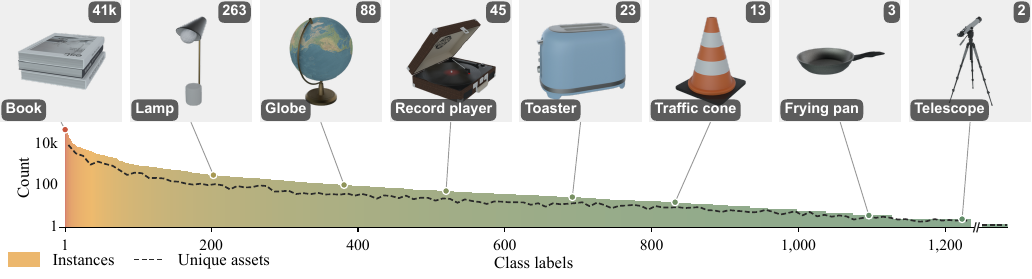}
  \caption{\textbf{Distribution of asset categories in \dataset.} The dataset contains over \assetsover high-quality assets across \classes asset classes.}
  \label{fig:asset-distribution}
\end{figure*}

\dataset contains 6,299 indoor scenes and 200 outdoor scenes.\footnote{\dataset is a curated subset of the Physicl scene library, which has since grown to over 10,000 scenes.}
Among the indoor scenes, 4,999 are household rooms and 1,300 are public indoor venues (\Cref{fig:scene-types} in the supplementary material). We split the scenes into 5,849 training, 325 validation, and 325 test scenes.
The scenes contain over \assetsover individual assets across \classes categories (\Cref{fig:asset-distribution}), and over \instances placements in total.
This shows that the dataset contains diverse assets and dense placements per scene (\Cref{tab:table_compare_artist}).

The scenes come with full PBR textures (albedo, roughness, and metallic) and explicit light emitters.
Each asset additionally carries per-part material annotations, physical properties, and metric scale.
This enables accurate physical simulation and photorealistic rendering of the scenes under different lighting conditions (\Cref{fig:scene-gallery}).

\subsection{Dataset creation}
\paragraph{Asset library.}
All assets and scenes in \dataset were created by professional 3D artists who were selected, trained, and equipped with dedicated tools. No content in this dataset is procedurally generated. Each object is modeled from real-world references and follows a common specification for geometry, metric scale, pivot placement and orientation, PBR materials, and naming. Before an object enters the library, it passes an automated technical validation and a human review of its visual quality, dimensions, and physical plausibility. Failed objects go back to their author until they pass.

\paragraph{Scene authoring.}
Artists compose the scenes from this library following real-world logic rather than procedural rules, and every scene passes an architectural and a technical review.
We then make each scene simulation-ready by computing collision hulls, deriving mass and inertia from the per-part material annotations, and defining explicit lighting from the artists' emitter materials (\Cref{sec:supp-creation} in the supplementary material).

\subsection{Physical interaction}
\label{sec:physical-interaction}
We use the simulation-ready scenes to build a mobile pick-and-place task with ground-truth grasps, affordances, and episodes.
The robot follows a natural-language instruction to navigate to the target object, pick it up with a feasible grasp, navigate to the target location, and place it, using only onboard camera images and proprioception, without maps or object poses.

\paragraph{Robot platform.}
We use a Franka Panda arm with a Franka Hand parallel gripper, mounted on a 3-DOF holonomic base.

\begin{figure}[t]
  \centering
  \includegraphics[width=\columnwidth]{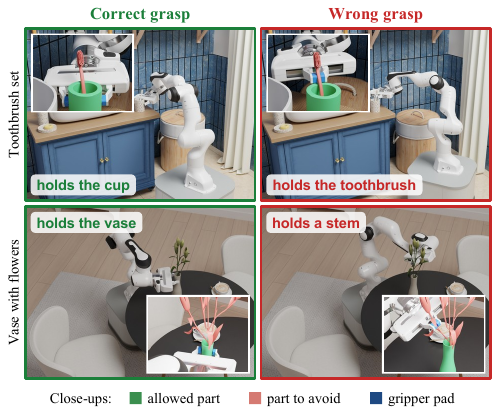}
  \caption{
    \textbf{Grasp affordance using part annotations of \dataset{}.}
    Left: grasps on the part the instruction implies.
    Right: successful picks that grasp the wrong part, penalized by affordance accuracy.}
  \label{fig:grasp-affordance}
\end{figure}

\begin{table*}[t]
  \centering
  \caption{\textbf{Cross-dataset generalization of single-view intrinsic decomposition.}
  We train IntrinsicImageDiffusion, Marigold-IID, and DNF-Intrinsic on InteriorVerse~\cite{zhu2022learning}, SAGE-10k~\cite{xia2026sage}, and \dataset, and evaluate them on datasets unseen during training.
  Int.\ is the intensity metric of MAW, a scale-invariant MSE of grayscale albedo.
  Training on \dataset improves cross-dataset generalization on most metrics.
  }
  \label{tab:ir-cross-dataset}
  \setlength{\tabcolsep}{4pt}
  \resizebox{\textwidth}{!}{%
  \begin{tabular}{@{}llcccccc@{}}
    \toprule
    & & \multicolumn{2}{c}{Hypersim~\cite{roberts2021hypersim}} & \multicolumn{3}{c}{Infinigen Indoors~\cite{raistrick2024infinigen}} & MAW~\cite{wu2023measured} \\
    \cmidrule(lr){3-4}\cmidrule(lr){5-7}\cmidrule(lr){8-8}
    Method & Train set & Alb.\ si-PSNR $\uparrow$ & Alb.\ LPIPS $\downarrow$ & Alb.\ si-PSNR $\uparrow$ & Rough.\ RMSE $\downarrow$ & Metal.\ RMSE $\downarrow$ & Int.\ $\downarrow$ \\
    \midrule
    \multirow{3}{*}{IntrinsicImageDiffusion~\cite{kocsis2024intrinsic}}
    & InteriorVerse~\cite{zhu2022learning} & 14.63 & 0.405 & 14.69 & 0.316 & 0.278 & 1.452 \\
    & SAGE-10k~\cite{xia2026sage} & 14.95 & 0.433 & \textbf{16.36} & 0.296 & \textbf{0.130} & 1.455 \\
    & \dataset (Ours)       & \textbf{15.24} & \textbf{0.392} & 16.14 & \textbf{0.268} & 0.172 & \textbf{1.319} \\
    \midrule
    \multirow{3}{*}{Marigold-IID~\cite{ke2025marigold}}
    & InteriorVerse~\cite{zhu2022learning} & 16.16 & 0.353 & 18.81 & 0.355 & 0.241 & 0.538 \\
    & SAGE-10k~\cite{xia2026sage} & 16.40 & \textbf{0.319} & 18.81 & 0.318 & 0.122 & 0.754 \\
    & \dataset (Ours)       & \textbf{17.31} & 0.323 & \textbf{21.06} & \textbf{0.252} & \textbf{0.109} & \textbf{0.322} \\
    \midrule
    \multirow{3}{*}{DNF-Intrinsic~\cite{zheng2025dnf}}
    & InteriorVerse~\cite{zhu2022learning} & 16.02 & 0.331 & 19.85 & 0.384 & 0.243 & 0.798 \\
    & SAGE-10k~\cite{xia2026sage} & 15.72 & 0.330 & 18.46 & 0.323 & 0.137 & 1.146 \\
    & \dataset (Ours)       & \textbf{16.85} & \textbf{0.316} & \textbf{21.40} & \textbf{0.250} & \textbf{0.122} & \textbf{0.423} \\
    \bottomrule
  \end{tabular}%
  }
\end{table*}

\paragraph{Grasp annotations.}
For each object, we sample candidate grasp poses with an antipodal grasp sampler for a parallel gripper. We keep the grasps that pass force-closure and collision checks, and verify their stability and success with simulated rollouts.
For some objects, grasping the wrong part should be avoided, \eg, a vase of flowers should be held by the vase rather than a stem (\Cref{fig:grasp-affordance}). Using the part annotations of \dataset, we label each grasp with the object part(s) supporting contact. This gives affordance labels for the 29\% of objects where the grasp location matters. For the other objects, such as a bottle, any stable grasp is acceptable.

\paragraph{Placement surfaces.}
Placing an object means putting it on a surface of another asset,
\eg, a table or shelf. For each such asset, we group the upward-facing faces of its mesh by height, which gives one surface per level, \eg, each board of an open shelf. A place target is a pose sampled on such a surface where the footprint of the object fits.

\paragraph{Navigation.}
To compute the base paths to the target object and to the place target, we rasterize the collision shapes of the scene into a 2D occupancy map and dilate it by the robot radius, so that every free cell is collision-free for the robot base. We then plan the shortest path over the free cells.

\paragraph{Expert demonstrations.}
An episode is one expert demonstration, recorded as a natural-language instruction, wrist and base camera views, and robot actions. Using the grasps, place targets, and navigation paths above, we plan the navigation and manipulation actions under the robot's kinematic and dynamic constraints, execute each plan as a closed-loop rollout in MuJoCo, and keep only the episodes that succeed. In total, we generate 30,084 episodes with 10M frames (139.9 hours of simulation) in a subset of 1,155 scenes and 2,052 objects, following the train/validation/test scene split of \dataset.

\section{Experiments}
We evaluate \dataset on four tasks: inverse rendering, 3D instance segmentation, 3D reconstruction, and mobile manipulation. For each task, we train state-of-the-art models on \dataset or on prior synthetic datasets, and compare how well they generalize to unseen data, including real-world datasets. For mobile manipulation, we fine-tune VLA models and evaluate them on held-out \dataset scenes and on real-to-sim environments. The supplementary material provides additional benchmarks on \dataset for inverse rendering and monocular depth and normal estimation.
\subsection{Inverse rendering}

\paragraph{Setup.} For each scene, we render 100 images with Blender Cycles from camera poses sampled in the walkable regions, together with albedo, roughness, metallic, normal, and depth maps. We randomize the lighting per frame (\eg, lamps on/off and day/night) and drop frames in which the camera is too close to the geometry or sees too little of the scene.

\begin{figure*}[t]
  \centering
  \includegraphics[width=\textwidth]{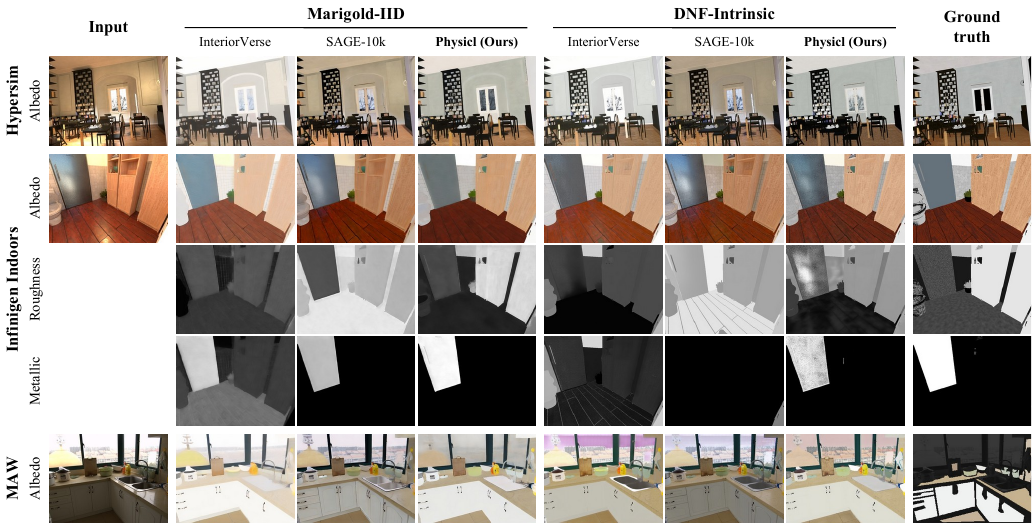}
  \caption{\textbf{Qualitative cross-dataset comparison of intrinsic decomposition.}
  Each method is trained on InteriorVerse~\cite{zhu2022learning}, SAGE-10k~\cite{xia2026sage}, or \dataset, and evaluated on datasets unseen during training: Hypersim~\cite{roberts2021hypersim}, Infinigen Indoors~\cite{raistrick2024infinigen}, and MAW~\cite{wu2023measured} (real photos).
  The model trained on \dataset reconstructs cleaner albedos (\eg, less lighting or shadow baked in) and gives roughness and metallic maps closer to the ground truth, even under challenging lighting.
  }
  \label{fig:ir-compare}
\end{figure*}

\paragraph{Baselines.} We train three diffusion-based intrinsic decomposition models, IntrinsicImageDiffusion~\cite{kocsis2024intrinsic}, Marigold-IID~\cite{ke2025marigold}, and DNF-Intrinsic~\cite{zheng2025dnf}, on \dataset and compare them with the same models trained on InteriorVerse~\cite{zhu2022learning}, a widely used dataset, and SAGE-10k~\cite{xia2026sage}, which is larger but only partly PBR-textured.

\paragraph{Evaluation.} We evaluate the models zero-shot on three unseen datasets: the synthetic Hypersim~\cite{roberts2021hypersim} (albedo si-PSNR and LPIPS) and Infinigen Indoors~\cite{raistrick2024infinigen} (albedo si-PSNR and roughness and metallic RMSE), and MAW~\cite{wu2023measured}, real photographs with measured albedo, where we report its intensity metric, the scale-invariant MSE of grayscale albedo over the annotated regions.

\paragraph{Results.} \Cref{tab:ir-cross-dataset} shows that models trained on \dataset{} generalize better than those trained on InteriorVerse or SAGE-10k, with the lowest roughness error on Infinigen Indoors and the lowest MAW intensity error for all three models. We attribute this to the diverse part materials and explicit scene lighting, which help models separate lighting from materials. In \Cref{fig:ir-compare}, the model trained on SAGE-10k bakes the dark lighting into the albedo, whereas the model trained on \dataset{} does not.

\subsection{3D instance segmentation}
\begin{table}[t]
  \centering
  \caption{
    \textbf{Zero-shot class-agnostic 3D instance segmentation.}
    Each method is trained on the GT instance masks of ScanNet200~\cite{rozenberszki2022language}, alone or joined with one synthetic set. Training on ScanNet200 and \dataset combined generalizes best on most metrics.}
  \label{tab:seg-class-agnostic}
  \setlength{\tabcolsep}{3.5pt}
  \small
  \resizebox{\ifdim\width>\columnwidth\columnwidth\else\width\fi}{!}{%
  \begin{tabular}{@{}lcccccc@{}}
    \toprule
    & \multicolumn{3}{c}{ScanNet++~\cite{yeshwanth2023scannet++}} & \multicolumn{3}{c}{Replica~\cite{straub2019replica}} \\
    \cmidrule(lr){2-4}\cmidrule(lr){5-7}
    Training data & AP $\uparrow$ & AP$_{50}$ $\uparrow$ & AP$_{25}$ $\uparrow$ & AP $\uparrow$ & AP$_{50}$ $\uparrow$ & AP$_{25}$ $\uparrow$ \\
    \midrule
    \multicolumn{7}{@{}l}{\textit{Mask3D}~\cite{schult2023mask3d}} \\
    ScanNet200 & 15.0 & 28.6 & 43.4 & 27.2 & 42.5 & 57.5 \\
    \quad + 3D-FRONT & 9.6 & 18.9 & 35.0 & 19.5 & 28.8 & 41.2 \\
    \quad + SAGE-10k & 18.9 & 34.1 & 47.5 & 32.3 & 50.9 & 64.6 \\
    \quad + \dataset (Ours) & \textbf{19.4} & \textbf{36.4} & \textbf{50.9} & \textbf{33.9} & \textbf{52.2} & \textbf{66.0} \\
    \midrule
    \multicolumn{7}{@{}l}{\textit{PTv3}~\cite{wu2024point}} \\
    ScanNet200 & 19.5 & 33.8 & 46.9 & 31.7 & 45.3 & 57.5 \\
    \quad + 3D-FRONT & 19.4 & 33.3 & 46.5 & 24.4 & 34.6 & 44.0 \\
    \quad + SAGE-10k & \textbf{22.2} & 38.1 & 50.5 & 30.8 & 47.5 & 60.5 \\
    \quad + \dataset (Ours) & 20.8 & \textbf{39.5} & \textbf{53.5} & \textbf{36.7} & \textbf{54.5} & \textbf{66.7} \\
    \bottomrule
  \end{tabular}%
  }
\end{table}

\paragraph{Setup.} To emulate real scans with occlusions and missing regions, we back-project the depth maps of rendered images into 3D point clouds for training. We also add ScanNet200~\cite{rozenberszki2022language} to the training set as a source of real-world data.

\paragraph{Baselines.} We train Mask3D~\cite{schult2023mask3d} and PTv3~\cite{wu2024point} for class-agnostic instance segmentation on the ground-truth instance masks of ScanNet200 combined with one synthetic dataset (3D-FRONT~\cite{fu20213d}, SAGE-10k~\cite{xia2026sage}, or \dataset). The supplementary material shows example training scans, open-vocabulary results, and qualitative examples.

\paragraph{Evaluation.} We evaluate zero-shot on ScanNet++~\cite{yeshwanth2023scannet++} and Replica~\cite{straub2019replica} and report AP, AP$_{50}$, and AP$_{25}$.

\paragraph{Results.}

\Cref{tab:seg-class-agnostic} shows that adding \dataset to ScanNet200 improves both models over ScanNet200 alone and beats the other synthetic datasets on all metrics but PTv3's ScanNet++ AP. Realistic and diverse synthetic scenes thus help 3D instance segmentation generalize, and the choice of synthetic dataset matters.

\subsection{3D reconstruction}

\begin{table*}[t]
  \centering
  \caption{\textbf{Zero-shot 3D reconstruction.}
    Each scene is reconstructed from 8 input views. We compute the 3D metrics for the GT observed parts. \dataset provides clean 3D supervision for reconstruction with accurate geometry, producing the lowest Chamfer distances.
    }
  \label{tab:recon-methods}
  \setlength{\tabcolsep}{4pt}
  \resizebox{\ifdim\width>\textwidth\textwidth\else\width\fi}{!}{%
  \begin{tabular}{@{}llcccccccc@{}}
    \toprule
    & & \multicolumn{4}{c}{ScanNet++~\cite{yeshwanth2023scannet++}} & \multicolumn{4}{c}{Replica~\cite{straub2019replica}} \\
    \cmidrule(lr){3-6}\cmidrule(lr){7-10}
    Method & Train set & Compl. 2D $\uparrow$ & AbsRel $\downarrow$ & Chamfer (cm) $\downarrow$ & F@10cm $\uparrow$ & Compl. 2D $\uparrow$ & AbsRel $\downarrow$ & Chamfer (cm) $\downarrow$ & F@10cm $\uparrow$ \\
    \midrule
    \multirow{3}{*}{FineRecon~\cite{stier2023finerecon}}
    & 3D-FRONT~\cite{fu20213d} & 0.732 & \textbf{0.125} & 13.62 & \textbf{0.744} & 0.640 & 0.148 & 19.88 & 0.653 \\
    & SAGE-10k~\cite{xia2026sage}    & 0.574 & 0.224 & 15.86 & 0.618 & 0.557 & \textbf{0.065} & 13.98 & 0.606 \\
    & \dataset (Ours)         & \textbf{0.758} & 0.137 & \textbf{12.43} & 0.712 & \textbf{0.860} & 0.084 & \textbf{8.46} & \textbf{0.727} \\
    \midrule
    \multirow{3}{*}{GenRecon~\cite{schmid2026genrecon}}
    & 3D-FRONT~\cite{fu20213d} & 0.972 & 0.118 & 9.96 & 0.751 & 0.974 & 0.049 & 7.99 & 0.727 \\
    & SAGE-10k~\cite{xia2026sage}    & 0.977 & 0.116 & 10.41 & 0.732 & \textbf{0.989} & 0.047 & 7.36 & 0.748 \\
    & \dataset (Ours)         & \textbf{0.983} & \textbf{0.096} & \textbf{8.85} & \textbf{0.801} & 0.986 & \textbf{0.039} & \textbf{6.47} & \textbf{0.824} \\
    \bottomrule
  \end{tabular}%
  }
\end{table*}

\begin{figure*}[t]
  \centering
  \includegraphics[width=0.9\textwidth]{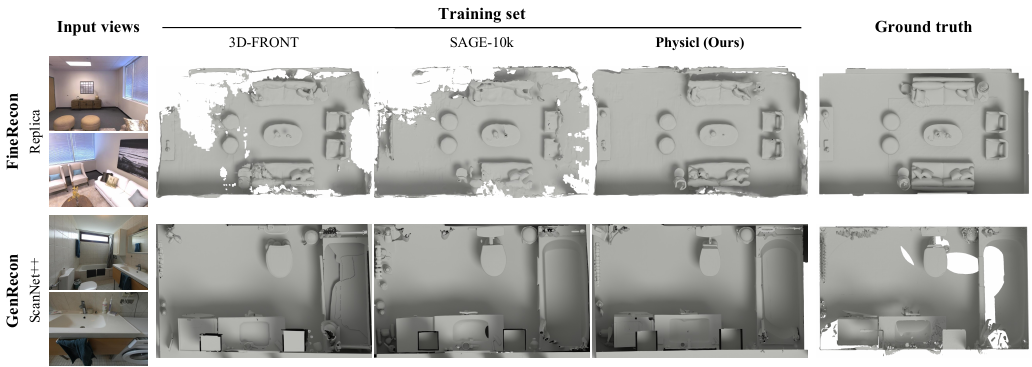}
  \caption{\textbf{Qualitative 3D reconstruction across training sets.}
  FineRecon on a Replica~\cite{straub2019replica} scene and GenRecon on a ScanNet++~\cite{yeshwanth2023scannet++} scene, each reconstructed from 8 input views, with the method trained on 3D-FRONT~\cite{fu20213d}, SAGE-10k~\cite{xia2026sage}, or \dataset. Our dataset yields sharper, more complete reconstructions.
  Predictions are clipped to the ground truth's floor plan, removing geometry outside the room.}
  \label{fig:recon-compare}
\end{figure*}

\paragraph{Setup.} Complete ground-truth geometry makes synthetic scenes well suited to train 3D reconstruction models. We render the input views with our pipeline and supervise with the ground-truth geometry and materials.

\paragraph{Baselines.} We train FineRecon~\cite{stier2023finerecon} and GenRecon~\cite{schmid2026genrecon} on 3D-FRONT~\cite{fu20213d}, SAGE-10k~\cite{xia2026sage}, or \dataset.

\paragraph{Evaluation.} We evaluate the models zero-shot on the real-world ScanNet++~\cite{yeshwanth2023scannet++} and Replica~\cite{straub2019replica} datasets from eight input views that uniformly cover each scene, ignoring regions that no input view sees.

\paragraph{Results.}
\Cref{tab:recon-methods} shows that training on \dataset gives the lowest Chamfer distance for both models on both datasets and the best results on most other metrics. We attribute this to the high-fidelity, artist-modeled geometry and assets of \dataset across \classes object categories (\Cref{tab:table_compare_artist,tab:table_compare_agent}), which give the models better shape priors and lead to sharper and more complete reconstructions (\Cref{fig:recon-compare}). In contrast, 3D-FRONT contains mostly large furniture.

\subsection{Mobile manipulation}
\paragraph{Setup.} We evaluate two settings. The first is our mobile pick-and-place benchmark (\Cref{sec:physical-interaction}) on the held-out \dataset test scenes (\pickplacetestscenes scenes and \pickplacetestepisodes episodes), where policies train on the demonstrations of the training scenes. The second tests zero-shot transfer to real-world scans with mobile pick-only on the ScanNet++ real-to-sim environments of SimNet++~\cite{simnet2026} (35 scenes and 183 episodes), where we compare VLAs trained on \dataset or on MolmoSpaces~\cite{kim2026molmospaces}.
\begin{table}[t]
  \centering
  \caption{
    \textbf{Mobile manipulation benchmark.} VLA models are fine-tuned on \dataset and evaluated on held-out test scenes. Even the best model, $\pi_{0.5}$, succeeds in only 15.2\% of the episodes, leaving large room for improvement. Time is averaged over successful episodes.
  }
  \label{tab:mobile-manip}
  \footnotesize
  \setlength{\tabcolsep}{3pt}
  \begin{tabular*}{\columnwidth}{@{}l@{\extracolsep{\fill}}cccccc@{}}
    \toprule
    Method & Reach & Pick & Surface & Success & Aff.\ acc. & Time (s) \\
    \midrule
    X-VLA~\cite{zheng2026x}            & 16.7 & 1.4 & 1.2 & 0.3 & 66.7 & 43.1 \\
    GR00T N1.7~\cite{bjorck2025gr00t}         & 60.1 & 14.1 & 10.2 & 7.7 & 58.2 & 45.3 \\
    $\pi_{0.5}$~\cite{intelligence2025pi_}      & 69.1 & \textbf{20.6} & \textbf{18.3} & \textbf{15.2} & \textbf{75.6} & 44.0 \\
    MolmoBot~\cite{deshpande2026molmob0t}      & \textbf{74.2} & 19.2 & 13.8 & 9.6 & 74.2 & 47.0 \\
    \bottomrule
  \end{tabular*}
\end{table}

\paragraph{Baselines.} We fine-tune state-of-the-art VLA models, X-VLA~\cite{zheng2026x}, GR00T N1.7~\cite{bjorck2025gr00t}, $\pi_{0.5}$~\cite{intelligence2025pi_}, and MolmoBot~\cite{deshpande2026molmob0t}, on the \dataset training episodes.

\paragraph{Evaluation.} The robot starts near the target object, and we report the success rate averaged over three seeds per episode, along with the partial goals Reach (the object), Pick, and Surface (reaching the placement surface). A grasp succeeds if the object is lifted 5\,cm and held for 2\,s. For objects with affordance labels, we also report affordance accuracy, the fraction of successful grasps that contact the intended part.
For pick and place, our scenes and episodes support both learned navigation policies and expert navigation planners. Popular VLAs such as $\pi_{0.5}$ and MolmoBot, however, are built for manipulation within a short base move of the object, not for long-distance navigation. Trained end-to-end on our full episodes, they never reached a distant placement goal (0\% success). We therefore evaluate a hybrid setup, following existing mobile manipulation benchmarks~\cite{puig2024habitat,yenamandra2023homerobot,shukla2025maniskill}: the policy moves the base and arm to pick and place the object, and our expert planner handles only the long-distance navigation in between.

\paragraph{Results.}
\providecommand{\cmarkc}{\textcolor{green!55!black}{\ding{51}}}
\providecommand{\xmarkc}{\textcolor{black!30}{\ding{55}}}
\begin{table}[t]
    \centering
    \small
    \setlength{\tabcolsep}{3pt}
    \caption{\textbf{Mobile manipulation transfer to real-to-sim ScanNet++~\cite{yeshwanth2023scannet++} scenes.} VLAs trained on \dataset outperform those trained on MolmoSpaces in terms of success rate, demonstrating better zero-shot transfer to more realistic environments.}
    \label{tab:mobile-manip-transfer}
    \begin{tabular*}{\linewidth}{@{}l@{\extracolsep{\fill}}lcc@{}}
      \toprule
      Method & Training set & Success rate & Time (s) \\
      \midrule
      \multirow{2}{*}{$\pi_{0.5}$~\cite{intelligence2025pi_}}   & MolmoSpaces~\cite{kim2026molmospaces}  & 5.6 & 12.1 \\
                                     & \dataset (Ours)  & \textbf{13.8} & 10.5 \\
      \midrule
      \multirow{2}{*}{MolmoBot~\cite{deshpande2026molmob0t}}  & MolmoSpaces~\cite{kim2026molmospaces}  & 9.5 & 10.9 \\
                                     & \dataset (Ours)  & \textbf{12.8} & 10.1 \\
      \bottomrule
    \end{tabular*}
\end{table}

\begin{table}[t]
    \centering
    \small
    \setlength{\tabcolsep}{3pt}
    \caption{
      \textbf{Ablation on training with affordance labels for mobile picking.}
      With affordance labels, the models can better predict feasible grasps and improve task success.
    }
    \label{tab:affordance-ablation}
    \begin{tabular*}{\linewidth}{@{}l@{\extracolsep{\fill}}ccc@{}}
      \toprule
      Method & Affordance labels & Success rate & Affordance acc. \\
      \midrule
      \multirow{2}{*}{$\pi_{0.5}$~\cite{intelligence2025pi_}}   & \xmarkc & 17.8 & 56.6 \\
                                     & \cmarkc & \textbf{19.4} & \textbf{65.3} \\
      \midrule
      \multirow{2}{*}{MolmoBot~\cite{deshpande2026molmob0t}}  & \xmarkc & 17.6 & 36.0 \\
                                     & \cmarkc & \textbf{23.0} & \textbf{49.5} \\
      \bottomrule
    \end{tabular*}
    \vspace{-6pt}
\end{table}

\begin{figure}[t]
  \centering
  \footnotesize
  \newcommand{\rolloutimg}[1]{\includegraphics[width=0.465\columnwidth]{images/manip_rollouts/#1.jpg}}
  \newcommand{\rolloutlabel}[1]{\rotatebox{90}{\makebox[0.349\columnwidth]{\textbf{#1}}}}
  \renewcommand{\arraystretch}{0}
  \begin{tabular}{@{}c@{\hspace{2pt}}c@{\hspace{2pt}}c@{}}
    & \parbox[b]{0.465\columnwidth}{\centering ``Go to the glass and pick it up.''\vspace{2pt}}
    & \parbox[b]{0.465\columnwidth}{\centering ``Go to the sandal and pick it up.''\vspace{2pt}} \\
    \rolloutlabel{Real}  & \rolloutimg{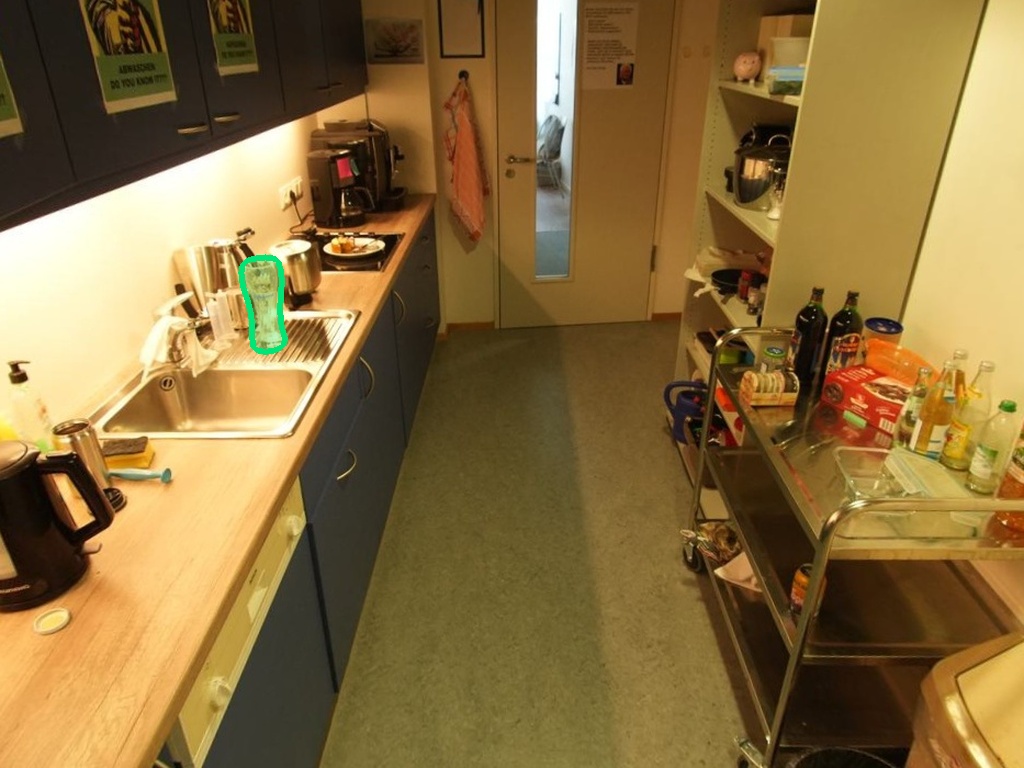}  & \rolloutimg{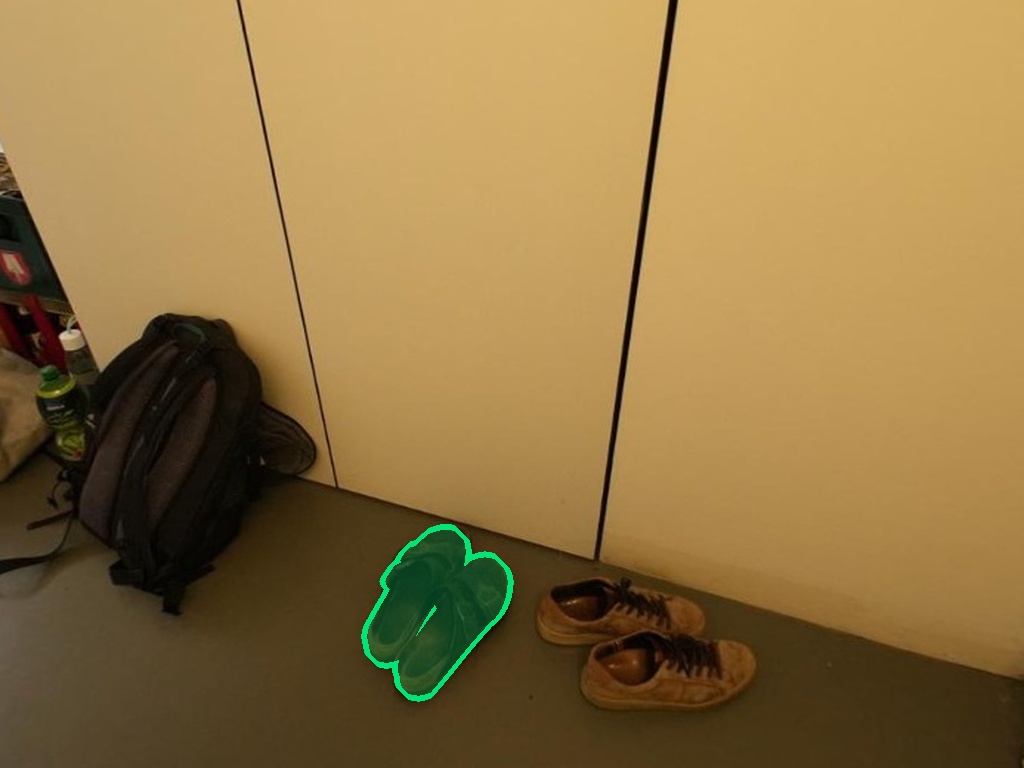}  \\[1.5pt]
    \rolloutlabel{Sim}   & \rolloutimg{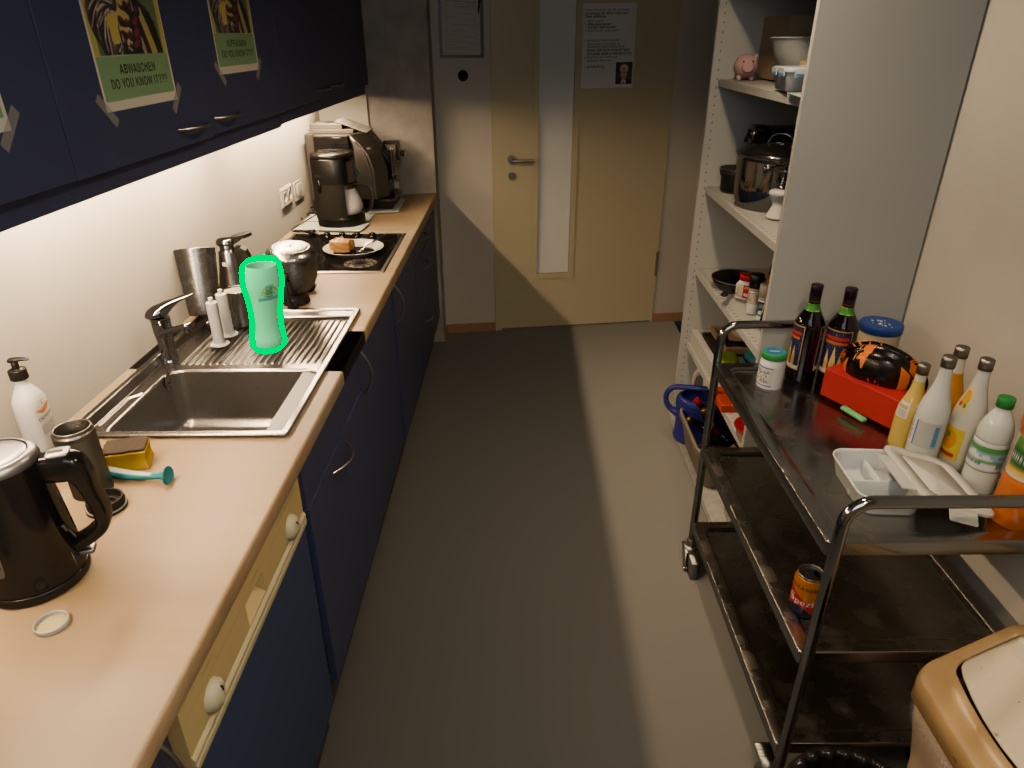}   & \rolloutimg{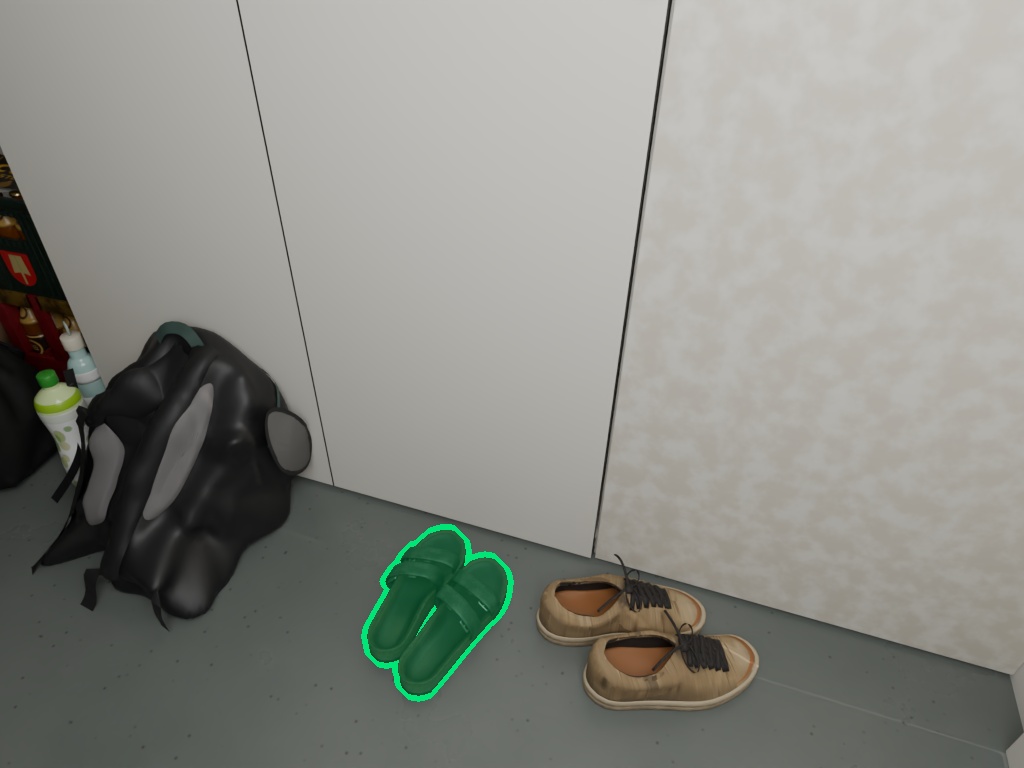}   \\[1.5pt]
    \rolloutlabel{Grasp} & \rolloutimg{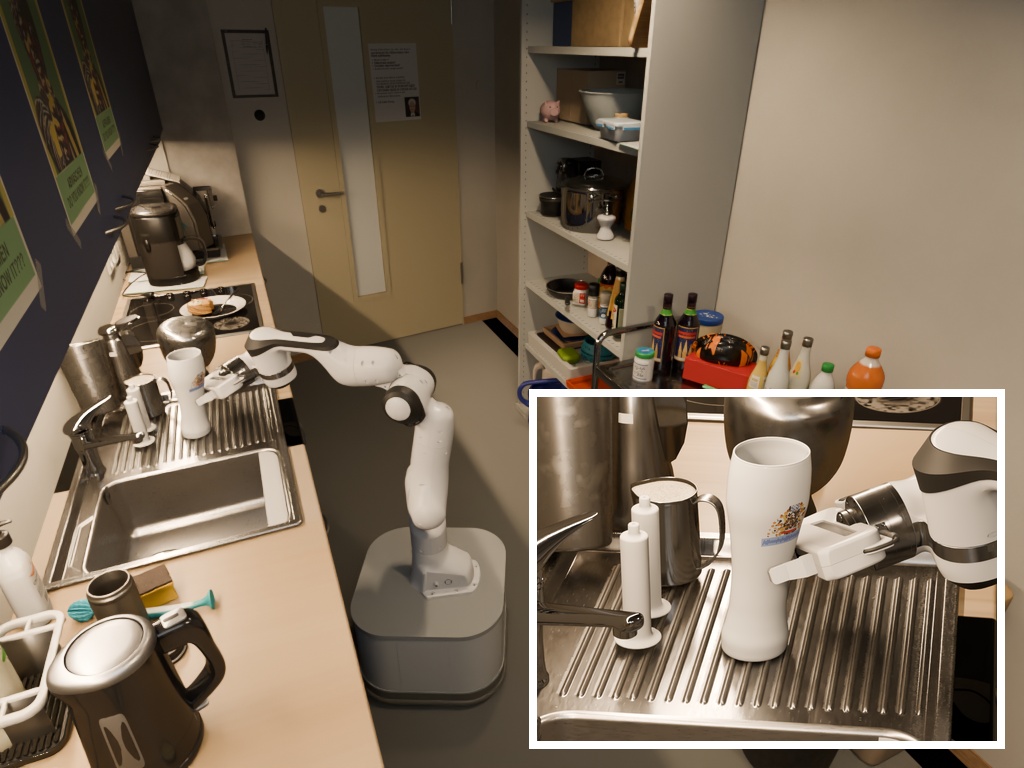} & \rolloutimg{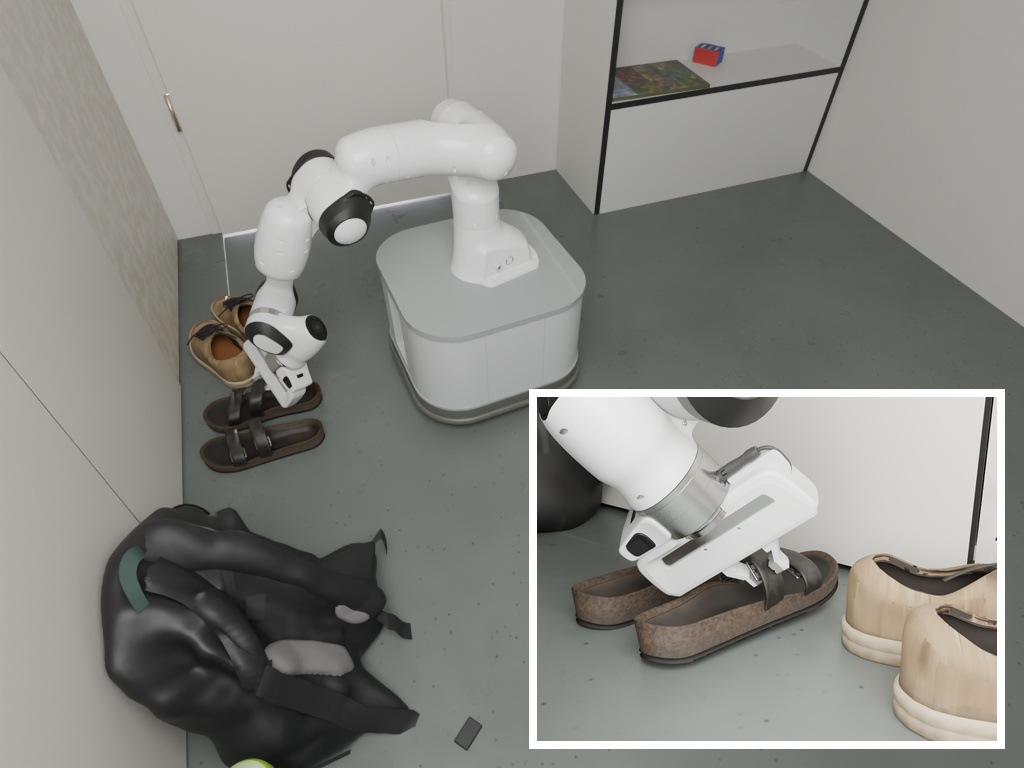} \\[1.5pt]
    \rolloutlabel{Lift}  & \rolloutimg{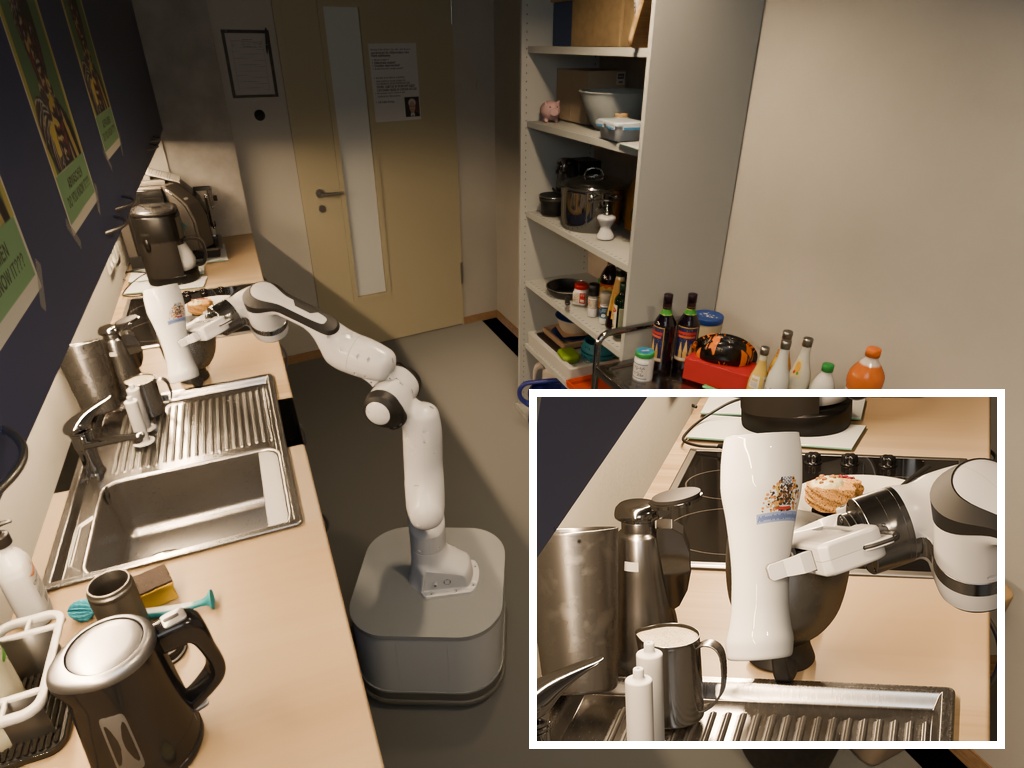}  & \rolloutimg{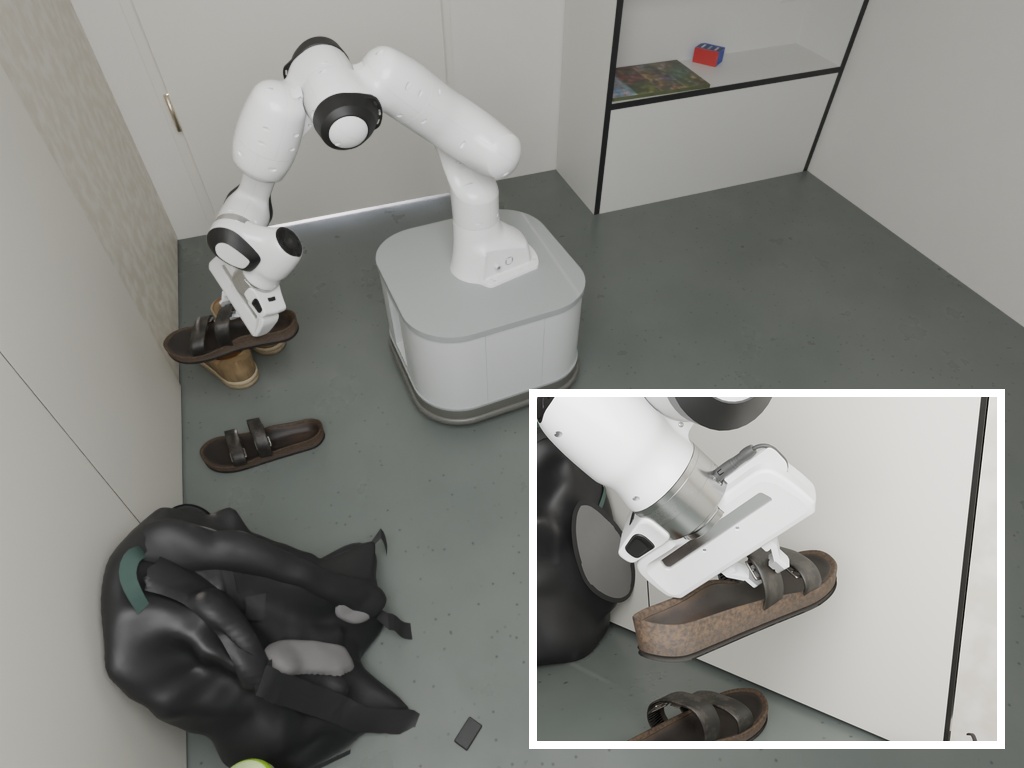}  \\
  \end{tabular}
  \caption{
    \textbf{Closed-loop mobile picking in ScanNet++ real-to-sim scenes.}
    Each column shows a real ScanNet++ photo and the simulated frames of a successful rollout by $\pi_{0.5}$ trained on \dataset, in a scene provided by SimNet++~\cite{simnet2026}.}
  \label{fig:manip-rollouts}
  \vspace{-6pt}
\end{figure}

\Cref{tab:mobile-manip} shows that although VLA models are effective on tabletop manipulation, they still struggle with long-horizon pick and place in complex, unseen scenes. In real-to-sim transfer, training on \dataset improves both VLAs over training on MolmoSpaces (\Cref{tab:mobile-manip-transfer} and \Cref{fig:manip-rollouts}).

\paragraph{Ablation.} \Cref{tab:affordance-ablation} compares $\pi_{0.5}$ and MolmoBot trained with and without affordance supervision, on picking the 61 objects with affordance labels (292 episodes). Affordance supervision improves both affordance accuracy and success rate, suggesting that it leads to more consistent grasps that are easier to execute.

\section{Limitations}
\label{sec:limitations}
\dataset is created and reviewed by artists, so it is more costly to extend than procedurally or agentically generated datasets.
Moreover, the scenes are static and all objects are simulated as rigid bodies, so they do not capture dynamics such as moving people, or deformable objects such as cloth.
For manipulation, our benchmark covers pick and place, but not tasks such as opening drawers or more abstract ones such as cleaning up a table.
Its affordance labels are also task independent, although the part to grasp can depend on the instruction, \eg, a mug is held by its handle to drink from but by its body to put away.

\section{Conclusion}

We presented \dataset, 6.5K artist-created, simulation-ready 3D scenes with full PBR materials, realistic layouts, and part-level annotations.
Training on \dataset improves generalization to unseen and real-world data for multiple tasks.
Our mobile manipulation benchmark evaluates action models on navigation and manipulation in diverse indoor scenes.
In the future, we hope \dataset will enable research that bridges perception and action in complex 3D environments, such as interactive 3D scene generation, action-conditioned world models, and embodied agents with spatial memory.

\ifdefined\BuildCameraReady
\section*{Acknowledgments}
We thank Physicl (\href{https://www.physicl.ai/}{\nolinkurl{physicl.ai}}), whose 3D artists created the scenes and assets in this dataset, for making them available for this work.
The scenes are drawn from the Physicl library, which has since grown to over 10,000 scenes.
Further renders and variations of these environments can be generated with Physicl Studio (\url{https://www.physicl.ai/studio}).
This work was supported by the ERC Starting Grant SpatialSem (101076253) and the ERC Consolidator Grant Gen3D (101171131).
The authors gratefully acknowledge the Gauss Centre for Supercomputing e.V. (\url{www.gauss-centre.eu}) for funding this project by providing computing time through the John von Neumann Institute for Computing (NIC) on the GCS Supercomputer JUWELS~\cite{JUWELS} at J{\"u}lich Supercomputing Centre (JSC).

\else
\clearpage
\fi
\section*{LLM Usage Statement}
We used large language models (LLMs) and vision-language models (VLMs) in the following parts of this work.
We used VLMs to classify each asset into a category and build the taxonomy. Qwen3.8-27B was the judge that scored layout realism, asset fidelity, and aesthetics in \Cref{tab:table_compare_artist,tab:table_compare_agent}. Claude Code assisted in writing the data processing, training, and evaluation code. The authors reviewed and tested all of it. Claude assisted in editing the text for grammar, wording, and clarity. The authors checked every edit and reference and take full responsibility for the content of the paper.

\printreferences
\ifdefined\BuildMainOnly\else

\fi
\fi

\end{document}